\documentclass[11pt,a4paper]{article}

\usepackage[top=0.75in, bottom=0.75in, left=0.85in, right=0.85in]{geometry}
\usepackage{amsmath,amssymb,amsfonts}
\usepackage{graphicx}
\usepackage{booktabs}
\usepackage{array}
\usepackage{multirow}
\usepackage{tabularx}
\usepackage{longtable}
\usepackage[dvipsnames,svgnames,table]{xcolor}
\usepackage{hyperref}
\usepackage{url}
\usepackage{natbib}
\usepackage{authblk}
\usepackage{abstract}
\usepackage{titlesec}
\usepackage{fancyhdr}
\usepackage{setspace}
\usepackage{microtype}
\usepackage{caption}
\usepackage{subcaption}
\usepackage{float}
\usepackage{enumitem}
\usepackage{algorithm}
\usepackage{algpseudocode}
\usepackage{listings}
\usepackage{tikz}
\usepackage{pgfplots}
\usepackage{placeins}
\pgfplotsset{compat=1.18}
\usepackage[utf8]{inputenc}
\usepackage{seqsplit}
\usepackage[htt]{hyphenat}

\definecolor{CodeBg}{RGB}{248,248,248}

\hypersetup{
  colorlinks=true,
  linkcolor=black,
  citecolor=black,
  urlcolor=blue,
  pdftitle={Detecting the Machine: A Comprehensive Benchmark of AI-Generated Text Detectors},
  pdfauthor={Madhav S. Baidya, S. S. Baidya, Chirag Chawla},
}

\titleformat{\section}{\large\bfseries}{\thesection}{1em}{}
\titleformat{\subsection}{\normalsize\bfseries}{\thesubsection}{1em}{}
\titleformat{\subsubsection}{\normalsize\bfseries\itshape}{\thesubsubsection}{1em}{}

\newcommand{\auroc}{\textsc{auroc}}
\newcommand{\auprc}{\textsc{auprc}}
\newcommand{\eer}{\textsc{eer}}

\newcommand{\hc}{\textsc{HC3}}
\newcommand{\eli}{\textsc{ELI5}}
\newcommand{\ELIFive}{\textsc{ELI5}}
\newcommand{\llm}{\textsc{llm}}
\newcommand{\CoT}{\textsc{CoT}}

\newcolumntype{C}[1]{>{\centering\arraybackslash}p{#1}}
\newcolumntype{L}[1]{>{\raggedright\arraybackslash}p{#1}}
\newcolumntype{R}[1]{>{\raggedleft\arraybackslash}p{#1}}

\begin{document}

\begin{titlepage}
\centering
\vspace*{1.5cm}

{\fontsize{20}{24}\selectfont\bfseries
Detecting the Machine:\\[6pt]
A Comprehensive Benchmark of AI-Generated\\[6pt]
Text Detectors Across Architectures,\\[6pt]
Domains, and Adversarial Conditions
\par}

\vspace{1.2cm}

{\large
\textbf{Madhav S. Baidya}$^{1}$,
\textbf{S.\,S. Baidya}$^{2}$,
\textbf{Chirag Chawla}$^{1}$ \par}

\vspace{0.6cm}

{\normalsize
$^{1}$Indian Institute of Technology (BHU), Varanasi, India \\
$^{2}$Indian Institute of Technology Guwahati, India \par}

\vspace{0.4cm}

{\small
\texttt{madhavsukla.baidya.chy22@itbhu.ac.in, saurav.baidya@iitg.ac.in,
chirag.chawla.chy22@itbhu.ac.in}
\par}

\vspace{0.8cm}
\hrule height 0.4pt
\vspace{0.8cm}

\begin{abstract}
\noindent
The rapid proliferation of large language models (\llm{}s) has created an urgent need for
robust, generalizable detectors of machine-generated text. Existing benchmarks typically
evaluate a single detector type on a single dataset under ideal conditions, leaving
critical questions about cross-domain transfer, cross-\llm{} generalization, and
adversarial robustness unanswered. This work presents a \textbf{comprehensive benchmark}
that systematically evaluates a broad spectrum of detection approaches across two carefully
constructed corpora: \hc{} (23{,}363 paired human--ChatGPT samples across five domains,
46{,}726 texts after binary expansion) and \eli{} (15{,}000 paired human--Mistral-7B
samples, 30{,}000 texts). The approaches evaluated span classical statistical classifiers,
five fine-tuned encoder transformers (BERT, RoBERTa, ELECTRA, DistilBERT, DeBERTa-v3),
a shallow 1D-CNN, a stylometric-hybrid XGBoost pipeline, perplexity-based unsupervised
detectors (GPT-2/GPT-Neo family), and \llm{}-as-detector prompting across four model
scales including GPT-4o-mini. All detectors are further evaluated zero-shot against
outputs from five unseen open-source \llm{}s with distributional shift analysis, and
subjected to iterative adversarial humanization at three rewriting intensities (L0--L2).
A principled length-matching preprocessing step is applied throughout to neutralize the
well-known length confound. Our central findings are: (\textit{i}) fine-tuned transformer
encoders achieve near-perfect in-distribution \auroc{} ($\geq\!0.994$) but degrade
universally under domain shift; (\textit{ii}) an XGBoost stylometric hybrid matches
transformer in-distribution performance while remaining fully interpretable,
with sentence-level perplexity coefficient of variation and AI-phrase density as the most
discriminative features; (\textit{iii}) \llm{}-as-detector prompting lags far behind
fine-tuned approaches --- the best open-source result is Llama-2-13b-chat-hf \CoT{} at \auroc{}
$0.898$, while GPT-4o-mini zero-shot reaches $0.909$ on \eli{} --- and is strongly
confounded by the generator--detector identity problem; (\textit{iv}) perplexity-based
detectors reveal a critical polarity inversion --- modern \llm{} outputs are
systematically \emph{lower} perplexity than human text --- that, once corrected, yields
effective \auroc{} of $\approx\!0.91$; and (\textit{v}) no detector generalizes
robustly across \llm{} sources and domains simultaneously. 
\end{abstract}

\vspace{0.6cm}
\hrule height 0.4pt
\vspace{0.4cm}


\vspace{0.4cm}
\hrule height 0.4pt
\end{titlepage}
\tableofcontents
\newpage

\section{Introduction}
\label{sec:intro}

The widespread deployment of instruction-tuned large language models --- including ChatGPT,
Mistral, LLaMA, and their successors \citep{brown2020gpt3, ouyang2022instructgpt,
bommasani2021foundation} --- has fundamentally altered the landscape of written
communication. These systems produce text that is, by many surface measures,
indistinguishable from human writing \citep{floridi2020gpt3}, giving rise to serious
societal concerns around academic integrity, journalistic authenticity, disinformation,
and the erosion of trust in digital communication. The development of robust, practical
detectors for machine-generated text has consequently become one of the most active
research frontiers in natural language processing \citep{mitchell2023detectgpt,
gehrmann2019gltr}.

Despite substantial progress, the field suffers from a critical methodological
fragmentation. Existing work evaluates detectors in isolation, on single datasets, under
idealized conditions that do not reflect the deployment environment. Key questions remain
empirically underexplored: \textit{How much does a detector's performance degrade when
the test-time \llm{} differs from the training-time generator? Which architectural
families generalize most robustly across domains? Can interpretable, lightweight detectors
match the performance of massive fine-tuned transformers? Does prompting large models with
structured reasoning constitute a viable detection strategy? What happens to all detector
families under adversarial text humanization?}

This paper addresses these questions through a large-scale, multi-stage benchmark that
spans the full spectrum of detection paradigms. Our contributions are:

To support reproducibility and further research, we make our implementation and evaluation
pipeline available at \href{https://github.com/MadsDoodle/Human-and-LLM-Generated-Text-Detectability-under-Adversarial-Humanization}{\texttt{our GitHub repository}}.

\begin{enumerate}[leftmargin=*,label=\textbf{\arabic*.}]
  \item \textbf{Benchmark design and datasets.} We construct two carefully controlled
        corpora --- \hc{} (paired human--ChatGPT, 5 domains, 46{,}726 samples after length
        matching) and \eli{} (paired human--Mistral-7B, single domain, 30{,}000 samples)
        --- with a principled length-matching step that prevents detectors from exploiting
        the length confound \citep{ippolito2020automatic}.

  \item \textbf{Three detector families (Stage 1).} We implement and rigorously evaluate
        under in-distribution and cross-domain conditions: (\textit{a}) classical
        statistical classifiers on a 22-feature hand-crafted feature set; (\textit{b})
        five fine-tuned encoder transformers --- BERT \citep{devlin2019bert}, RoBERTa
        \citep{liu2019roberta}, ELECTRA \citep{clark2020electra}, DistilBERT
        \citep{sanh2019distilbert}, DeBERTa-v3 \citep{he2021deberta}; (\textit{c}) a
        shallow 1D-CNN \citep{kim2014cnn}; (\textit{d}) a stylometric-hybrid XGBoost
        \citep{chen2016xgboost} pipeline with 60+ features including sentence-level
        perplexity and AI-phrase density; (\textit{e}) perplexity-based unsupervised
        detectors (GPT-2/GPT-Neo family); and (\textit{f}) \llm{}-as-detector prompting
        across four model scales (1.1B--14B parameters) including GPT-4o-mini via the
        OpenAI API.

  \item \textbf{Cross-\llm{} generalization (Stage 2).} All Stage 1 detectors are
        evaluated zero-shot against outputs from five unseen open-source \llm{}s
        (TinyLlama-1.1B, Qwen2.5-1.5B, Qwen2.5-7B,Llama-3.1-8B-Instruct, LLaMA-2-13B),
        complemented by embedding-space generalization via classical classifiers and
        distributional shift analysis in DeBERTa representation space.

  \item \textbf{Adversarial humanization (Stage 3).} All detectors are evaluated under
        three levels of iterative LLM-based rewriting (L0: original, L1: light
        humanization, L2: heavy humanization) using Qwen2.5-1.5B-Instruct as the
        rewriting model, probing robustness to the most practical evasion strategy
        available to adversarial users.

\end{enumerate}

\begin{figure}[H]
\centering
\includegraphics[width=0.78\textwidth]{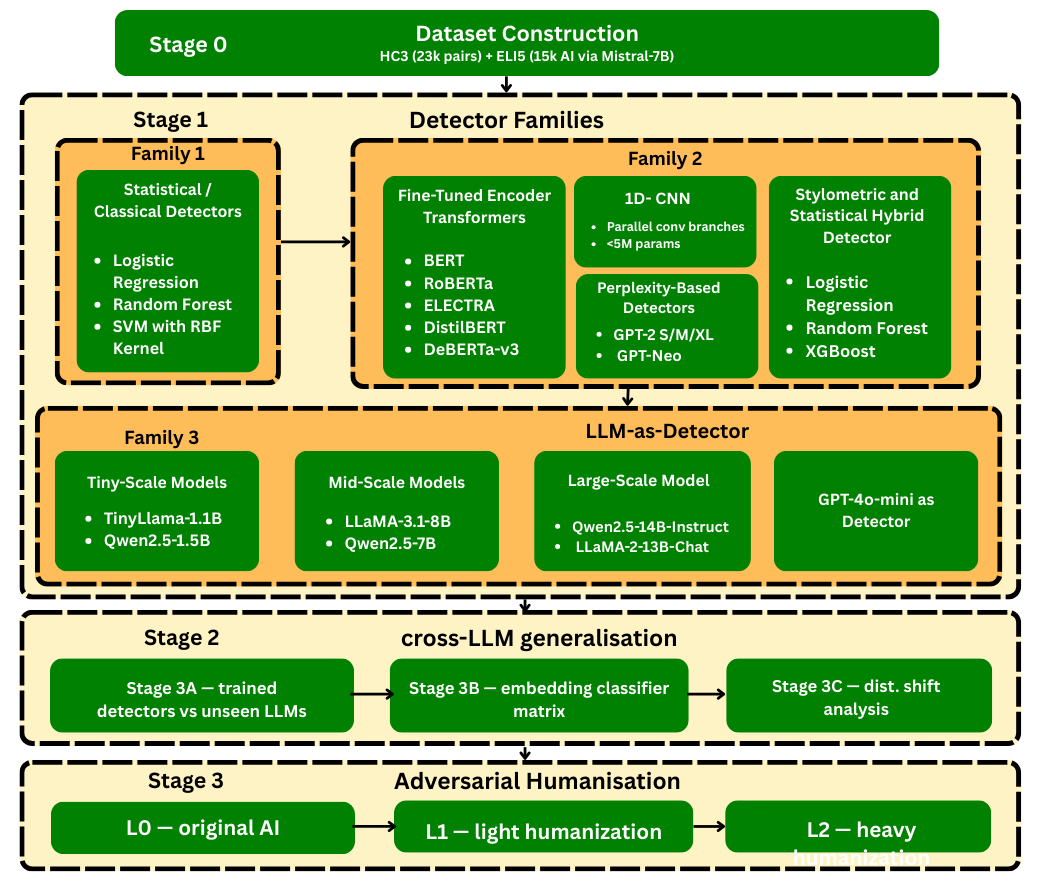}
\caption{Overview of the benchmark pipeline. \textbf{Stage 0} constructs two paired
corpora (\hc{}: 23k human--ChatGPT pairs; \eli{}: 15k human--Mistral-7B pairs) with
length-matched preprocessing. \textbf{Stage 1} evaluates three detector families:
Family~1 (classical statistical classifiers), and Family~2 (fine-tuned encoder
transformers --- BERT, RoBERTa, ELECTRA, DistilBERT, DeBERTa-v3; 1D-CNN;
perplexity-based detectors; stylometric-hybrid XGBoost), and Family~3
(\llm{}-as-detector prompting at four scales including GPT-4o-mini). \textbf{Stage~2}
evaluates cross-\llm{} generalization via neural detectors, embedding-space
classifier matrices , and distributional shift analysis. \textbf{Stage~3}
applies adversarial humanization at three levels (L0--L2) using an instruction-tuned
rewriter. All families are evaluated under a unified five-metric suite (\auroc{},
\auprc{}, \eer{}, Brier Score, FPR@95\%TPR).}
\label{fig:architecture}
\end{figure}

\section{Related Work}
\label{sec:related}

\subsection{Supervised Detection Approaches}

Early work on machine-generated text detection relied on statistical features such as
perplexity under reference language models \citep{solaiman2019release}, n-gram statistics,
and stylometric signals \citep{juola2006authorship, stamatatos2009survey,
ippolito2020automatic}. The introduction of transformer-based detectors substantially
advanced the field: models such as GROVER \citep{zellers2019defending} demonstrated that
the best generators also serve as the best discriminators. Subsequent work fine-tuned
general-purpose encoders (BERT, RoBERTa) on paired human/\llm{} corpora, achieving high
in-distribution accuracy \citep{rodriguez2022cross}. The \hc{} corpus \citep{guo2023close}
introduced a systematic multi-domain benchmark for ChatGPT detection that has become a
de-facto standard. Several subsequent studies have investigated domain transfer
\citep{uchendu2020authorship}, adversarial robustness \citep{wolff2022attacking}, and
the effect of prompt engineering on detectability. Commercial detection tools have also
been deployed \citep{openai2023classifier}, though their generalization across \llm{}
families remains poorly characterized.

\subsection{Unsupervised and Zero-Shot Approaches}

DetectGPT \citep{mitchell2023detectgpt} exploits the observation that \llm{}-generated
text tends to lie in local probability maxima of the generating model, using
perturbation-based curvature estimation as a detection signal. Statistical visualization
tools such as GLTR \citep{gehrmann2019gltr} provide complementary token-level detection
signals. Perplexity thresholding under reference models has been widely studied
\citep{lavergne2008detecting}, though as we show, the direction of the perplexity signal
is counter-intuitive in the modern \llm{} era. Watermarking schemes
\citep{kirchenbauer2023watermark} provide a complementary but generator-controlled
approach that requires cooperation from the model provider.

\subsection{LLM-as-Detector}

The use of large models as zero-shot or few-shot classifiers for their own outputs has
been explored in several recent studies \citep{zeng2023evaluating, bhattacharjee2023conda}.
A consistent finding is that prompting-based detection underperforms fine-tuned approaches,
particularly on out-of-distribution text. Chain-of-thought prompting has been shown to
improve classification accuracy for models with sufficient instruction-following capacity
\citep{kojima2022large, wei2022chain}, a finding we confirm and extend across four
model scales.

\subsection{Adversarial Humanization}

Paraphrase-based attacks \citep{krishna2023paraphrasing}, style transfer, and direct
human editing have all been demonstrated to substantially reduce detector accuracy. The
challenge of adversarial robustness remains largely unsolved, particularly for unsupervised
detection methods. Our Stage 2 evaluation systematically characterizes how iterative
\llm{}-based rewriting at two intensity levels degrades all detector families simultaneously,
filling a gap left by prior work that typically evaluates a single detector family under
a single attack strategy.
\section{Datasets and Preprocessing}
\label{sec:data}

\subsection{HC3 Dataset}

The \hc{} (Human-ChatGPT Comparison) corpus \citep{guo2023close} was loaded from the
\texttt{Hello-SimpleAI/HC3} repository via the Hugging Face \texttt{datasets} library.  It
provides question--answer pairs across multiple domains, with each entry containing one
question, a list of human answers, and a list of ChatGPT answers.  We flattened the corpus
into a structured paired format --- one row per question with a single human answer and a
single ChatGPT answer --- yielding 47{,}734 paired examples across six domain splits
(Table~\ref{tab:hc3_domains}).  Following exact-duplicate removal on the question field,
the corpus was reduced to 23{,}363 unique records.

\begin{table}[H]
\centering
\caption{Domain distribution of the \hc{} corpus after flattening and deduplication.}
\label{tab:hc3_domains}
\small
\begin{tabular}{lc}
\toprule
\textbf{Domain} & \textbf{Unique Pairs} \\
\midrule
\texttt{reddit\_eli5} & 16{,}153 \\
\texttt{finance}      &  3{,}933 \\
\texttt{medicine}     &  1{,}248 \\
\texttt{open\_qa}     &  1{,}187 \\
\texttt{wiki\_csai}   &    842 \\
\midrule
\textbf{Total}        & 23{,}363 \\
\bottomrule
\end{tabular}
\end{table}

\subsection{\eli{} Dataset and Mistral-7B Augmentation}

The \eli{} dataset \citep{fan2019eli5} was loaded from \texttt{sentence-transformers/eli5}
via the Hugging Face hub.  It is a human-only question-answering corpus sourced from the
Reddit community \texttt{r/explainlikeimfive}, containing 325{,}475 training samples with
plain-language explanations of complex topics.  No \llm{}-generated answers exist in the
raw \eli{} data.

To create a balanced human--\llm{} paired corpus, we used \textbf{Mistral-7B-Instruct-v0.2}
to generate AI answers for a random sample of 15{,}000 \eli{} questions.  The generation
pipeline was optimized for throughput on an NVIDIA A100 GPU (Table~\ref{tab:mistral_config}).

\begin{table}[H]
\centering
\caption{Mistral-7B generation configuration for \eli{} augmentation.}
\label{tab:mistral_config}
\small
\begin{tabular}{ll}
\toprule
\textbf{Parameter} & \textbf{Value} \\
\midrule
Model              & \texttt{mistralai/Mistral-7B-Instruct-v0.2} \\
Precision          & FP16 (no quantization) \\
Attention          & Flash Attention 2 \\
Compilation        & \texttt{torch.compile} (reduce-overhead) \\
Batch size         & 48 \\
Max new tokens     & 150 \\
Temperature        & 0.7 \\
Top-$p$            & 0.9 \\
\bottomrule
\end{tabular}
\end{table}

Each question was formatted using Mistral's \texttt{[INST]} instruction template and fed
through the model in batches.  Generated tokens were decoded with the prompt stripped,
yielding clean answer strings.

\subsection{Binary Dataset Preparation and Length Matching}
\label{sec:length_matching}

AI-generated text detection is formulated as a binary classification problem.  Rather than
treating each question--answer pair as a unit, every individual answer is treated as an
independent text sample labeled either \textit{human} (0) or \textit{llm} (1).  This
decoupling reflects the actual deployment setting, where detectors receive isolated text
snippets with no access to the corresponding question.

This conversion yielded perfectly balanced binary corpora:
\begin{itemize}[leftmargin=*]
  \item \textbf{\hc{} binary}: 46{,}726 samples (23{,}363 human + 23{,}363 \llm{})
  \item \textbf{\eli{} binary}: 30{,}000 samples (15{,}000 human + 15{,}000 \llm{})
\end{itemize}

A critical \textbf{length-matching} step was applied before splitting.  It is well-documented that \llm{}-generated answers are systematically longer than human
answers \citep{ippolito2020automatic}; without correction,
a classifier can achieve high accuracy by learning response length --- a spurious, non-linguistic
shortcut that collapses under any length-normalized adversarial condition.  Each human answer
was therefore matched with an \llm{} answer falling within $\pm 20\%$ of its word count,
ensuring statistically comparable length distributions across classes.

Stratified 80/20 train/test splits were then constructed, preserving exact class balance
(Table~\ref{tab:splits}).

\begin{table}[H]
\centering
\caption{Train/test split sizes after length matching and stratification.}
\label{tab:splits}
\small
\begin{tabular}{lcc}
\toprule
\textbf{Dataset} & \textbf{Train} & \textbf{Test} \\
\midrule
\hc{}  & 36{,}968 (18{,}484 per class) & 9{,}242 \\
\eli{} & 22{,}862 (11{,}431 per class) & 5{,}716 \\
\bottomrule
\end{tabular}
\end{table}

The two datasets are kept separate throughout all evaluations: \hc{} represents a formal,
multi-domain corpus with ChatGPT as the \llm{} source, while \eli{} represents a
conversational, single-domain corpus with Mistral-7B as the source.  This separation
enables cross-dataset generalization analysis.

\section{Detector Families: Architecture and Implementation}
\label{sec:detectors}

All detectors output a continuous detectability score in $[0, 1]$ representing the
probability that a given text is \llm{}-generated.  Each supervised family is trained and
evaluated under four conditions: in-distribution (same dataset for train and test) and
cross-distribution (train on one dataset, test on the other), producing a $2\!\times\!2$
evaluation grid per detector.  Unsupervised and zero-parameter families are evaluated on
both test sets without training.

\subsection{Statistical / Classical Detectors}
\label{sec:classical}

This family operates entirely on hand-crafted linguistic features with no learned
representations.  The feature extractor computes \textbf{22 interpretable signals} organized
into seven categories:

\begin{enumerate}[leftmargin=*,label=\textit{(\roman*)}]
  \item \textbf{Surface statistics}: word count, character count, sentence count, average
        word length, average sentence length.
  \item \textbf{Lexical diversity}: type-token ratio, hapax legomena ratio.
  \item \textbf{Punctuation and formatting}: comma density, period density, question mark
        ratio, exclamation ratio.
  \item \textbf{Repetition metrics}: bigram repetition rate, trigram repetition rate.
  \item \textbf{Entropy measures}: Shannon entropy over word-frequency distribution,
        sentence-length entropy.
  \item \textbf{Syntactic complexity}: sentence-length variance and standard deviation.
  \item \textbf{Discourse markers}: hedging-word density, certainty-word density,
        connector-word density, contraction ratio, burstiness.
\end{enumerate}

Three classifiers are trained on this feature vector:
\textbf{Logistic Regression} (L2 penalty, interpretable linear baseline);
\textbf{Random Forest} (100 trees, max depth 10, bootstrap sampling); and
\textbf{SVM with RBF Kernel} (Platt-scaled probabilities).

\subsection{Fine-Tuned Encoder Transformers}
\label{sec:transformers}

All transformer models share a common fine-tuning protocol: the pre-trained encoder is
loaded with a two-class classification head appended to the \texttt{[CLS]} token
representation, then fine-tuned end-to-end for one epoch on binary human/\llm{} labels.

Training uses AdamW ($\text{lr} = 2{\times}10^{-5}$, weight decay $= 0.01$), linear
warmup over 6\% of training steps, dropout increased to 0.2, a 10\% held-out validation
split for early stopping (patience $= 3$), and \auroc{} as the model-selection criterion.

Mixed precision (FP16) is used throughout.  Batch size is 32 (train) and 64 (eval).
The detectability score is the softmax probability assigned to the \llm{} class.

\subsubsection{BERT (\texttt{bert-base-uncased})}

BERT \citep{devlin2019bert} uses bidirectional masked language modeling pre-training,
processing full token sequences with attention over both left and right context.
The base variant has 12 transformer layers, 12 attention heads, hidden size 768,
intermediate size 3{,}072, and $\approx\!110$M parameters.  Tokenization uses WordPiece
with a 30{,}522-token vocabulary; sequences are truncated to 512 tokens.

\subsubsection{RoBERTa (\texttt{roberta-base})}

RoBERTa \citep{liu2019roberta} improves upon BERT by removing the next-sentence
prediction objective, training on $10\times$ more data with larger batches, using dynamic
masking, and employing a Byte-Pair Encoding tokenizer (50{,}265-token vocabulary).
It shares the same 12-layer, 768-hidden, 125M parameter architecture but benefits from
more robust pre-training.

\subsubsection{ELECTRA (\texttt{google/electra-base-discriminator})}

ELECTRA \citep{clark2020electra} replaces masked language modeling with a replaced-token
detection objective: a small generator corrupts tokens and the discriminator is trained to
identify which tokens were replaced.  This produces more sample-efficient pre-training,
as every token position contributes a training signal (vs.\ $\approx\!15\%$ in BERT).
ELECTRA's token-level discriminative pre-training makes it particularly sensitive to
local stylistic anomalies common in \llm{} outputs.

\subsubsection{DistilBERT (\texttt{distilbert-base-uncased})}

DistilBERT \citep{sanh2019distilbert} is a knowledge-distilled compression of BERT,
retaining 97\% of BERT's language understanding at 60\% of its parameter count
($\approx\!66$M parameters, 6 layers).  Distillation uses a soft-label cross-entropy loss
against the teacher BERT's output distribution, combined with cosine embedding alignment.
DistilBERT is particularly attractive for deployment-scale detection systems due to its
significantly reduced inference latency.

\subsubsection{DeBERTa-v3 (\texttt{microsoft/deberta-v3-base})}

DeBERTa-v3 \citep{he2021deberta} introduces two architectural advances over RoBERTa.
First, \textit{disentangled attention}: each token is represented by two separate vectors
--- one for content and one for relative position --- with attention weights computed across
all four content-position cross-interactions.  Second, DeBERTa-v3 adopts ELECTRA-style
replaced token detection for pre-training rather than masked language modelling.  The base
model has approximately 184M parameters.

\textbf{Implementation.}  A critical consideration for DeBERTa-v3 is precision handling.
The disentangled attention mechanism produces gradient magnitudes that underflow in BF16's
7-bit mantissa, rendering mixed-precision training numerically unsafe for this architecture.
Training is therefore conducted in full FP32 throughout (\texttt{fp16=False},
\texttt{bf16=False}, with an explicit \texttt{model.float()} cast at initialization).
Checkpoint reloading is disabled entirely (\texttt{save\_strategy="no"},
\texttt{load\_best\_model\_at\_end=False}), and final in-memory weights are used directly
for prediction --- this avoids the LayerNorm parameter naming inconsistency between saved
and reloaded checkpoints that is a known fragility of DeBERTa-v3 under the HuggingFace
Trainer.  Explicit gradient clipping (\texttt{max\_grad\_norm=1.0}) is applied for
training stability.  \texttt{token\_type\_ids} are intentionally omitted, as DeBERTa-v3
does not use segment IDs.  DeBERTa-v3 uses AdamW ($\text{lr}=2\times10^{-5}$,
weight\_decay$=0.01$), 500 warmup steps (fixed, not ratio-based), 1 epoch, batch size 16.

\subsection{Shallow 1D-CNN Detector}
\label{sec:cnn}

The 1D-CNN detector is a lightweight neural model targeting local $n$-gram patterns rather
than global sequence context, following the architecture of \citet{kim2014cnn}.  It follows the architecture:
\[
\text{Embedding} \to \text{Parallel 1D-Conv} \to \text{Global Max Pool}
\to \text{Dense Head} \to \sigma
\]
A shared embedding layer (vocab size 30{,}000, dim 128) projects token IDs into dense
vectors.  Four parallel convolutional branches with kernel sizes $\{2,3,4,5\}$ each produce
128 feature maps (BatchNorm + ReLU).  Global max pooling extracts the most salient
activation per filter, producing a 512-dimensional concatenated feature vector.  A
two-layer dense head ($512 \to 256 \to 1$) with dropout (0.4) and sigmoid output produces
the detectability score.

Texts are truncated to 256 tokens (shorter than the transformer maximum of 512, as local
$n$-gram patterns are captured in shorter windows).  Total parameter count is under 5M ---
intentionally constrained to probe whether shallow learned representations can bridge the
gap between handcrafted features and full transformer fine-tuning.

Training uses Adam ($\text{lr} = 10^{-3}$), ReduceLROnPlateau scheduling (factor 0.5,
patience 1), gradient clipping (norm 1.0), and early stopping (patience $= 3$) over up
to 10 epochs.

\subsection{Stylometric and Statistical Hybrid Detector}
\label{sec:stylometric}

This family substantially extends the classical feature set from 22 to 60+ features across
eight categories, adding:

\begin{itemize}[leftmargin=*]
  \item \textbf{AI phrase density}: frequency of structurally AI-characteristic phrases
        (\textit{e.g.}, ``it is worth noting'', ``in summary'', ``to summarize'').
  \item \textbf{Function word frequency profiles}: overall function word ratio plus
        per-word frequency for the 10 most common function words.
  \item \textbf{Punctuation entropy}: Shannon entropy over the punctuation character
        distribution --- \llm{} text tends toward lower entropy (more uniform punctuation).
  \item \textbf{Readability indices}: Flesch Reading Ease, Flesch-Kincaid Grade,
        Gunning Fog, SMOG Index, ARI, Coleman-Liau Index.
  \item \textbf{POS tag distribution} (spaCy): normalized frequency of 10 POS categories.
  \item \textbf{Dependency tree depth}: mean and maximum parse-tree depth across sentences.
  \item \textbf{Sentence-level perplexity} (GPT-2 Small): mean, variance, standard
        deviation, and \textbf{coefficient of variation (CV)} of per-sentence perplexity.
        The CV is particularly diagnostic: \llm{} text exhibits uniformly low perplexity
        (low CV), while human text varies considerably across sentences (high CV).
\end{itemize}

Three classifiers are trained: Logistic Regression (L2, lbfgs solver), Random Forest
(300 trees, max depth 12), and \textbf{XGBoost} \citep{chen2016xgboost} (400 estimators, learning rate 0.05,
depth 6, subsample 0.8).  All features are standardized via \texttt{StandardScaler}.

\subsection{Perplexity-Based Detectors}
\label{sec:ppl_arch}

Perplexity-based detection is an unsupervised, training-free approach that exploits the
distributional overlap between autoregressive reference models and \llm{}-generated text.
Because GPT-2 and GPT-Neo family models share training corpus overlap with modern \llm{}
generators, they assign systematically \emph{lower} perplexity to \llm{}-generated text
than to human-written text.  The detectability score is therefore an \textbf{inversion}
of the raw perplexity signal.  Five reference models are evaluated: GPT-2 Small (117M),
GPT-2 Medium (345M), GPT-2 XL (1.5B), GPT-Neo-125M, and GPT-Neo-1.3B.  Full
implementation details and the sliding-window strategy for long texts are described in
Section~\ref{sec:perplexity}.

\subsection{LLM-as-Detector}
\label{sec:llm_arch}

The \llm{}-as-detector paradigm treats generative language models as zero-parameter
classifiers, deriving detectability scores from constrained decoding logits (for local
models) or structured rubric scores (for API models).  Five open-source models spanning
1.1B to 14B parameters are evaluated (TinyLlama-1.1B, Qwen2.5-1.5B, Qwen2.5-7B,
LLaMA-3.1-8B, LLaMA-2-13B-Chat), along with GPT-4o-mini via the OpenAI API.  Full
implementation details including prompt polarity correction, task prior subtraction,
and the hybrid confidence-logit scoring scheme are described in
Section~\ref{sec:llm_detector}.

\section{Experimental Results: Detector Families}
\label{sec:results1}

\subsection{Statistical / Classical Detectors}

Tables~\ref{tab:lr_results}--\ref{tab:svm_results} report results for Logistic Regression,
Random Forest, and SVM with RBF kernel.

\begin{table}[H]
\centering
\caption{Logistic Regression results across evaluation conditions.}
\label{tab:lr_results}
\small
\begin{tabular}{lrrrrr}
\toprule
\textbf{Condition} & \textbf{\auroc{}} & \textbf{Brier} & \textbf{Log Loss} & \textbf{Mean Human} & \textbf{Mean \llm{}} \\
\midrule
hc3\_to\_hc3   & 0.8882 & 0.1334 & 0.4411 & 0.2838 & 0.7319 \\
hc3\_to\_eli5  & 0.7406 & 0.2116 & 0.6474 & 0.4246 & 0.6508 \\
eli5\_to\_eli5 & 0.8446 & 0.1605 & 0.4909 & 0.3251 & 0.6760 \\
eli5\_to\_hc3  & 0.7429 & 0.2496 & 0.9063 & 0.2006 & 0.4580 \\
\bottomrule
\end{tabular}
\end{table}

\begin{table}[H]
\centering
\caption{Random Forest results across evaluation conditions.}
\label{tab:rf_results}
\small
\begin{tabular}{lrrrrr}
\toprule
\textbf{Condition} & \textbf{\auroc{}} & \textbf{Brier} & \textbf{Log Loss} & \textbf{Mean Human} & \textbf{Mean \llm{}} \\
\midrule
hc3\_to\_hc3   & 0.9767 & 0.0679 & 0.2438 & 0.1889 & 0.8173 \\
hc3\_to\_eli5  & 0.7829 & 0.1922 & 0.5815 & 0.3830 & 0.6086 \\
eli5\_to\_eli5 & 0.9618 & 0.0869 & 0.3014 & 0.2348 & 0.7811 \\
eli5\_to\_hc3  & 0.6337 & 0.3193 & 1.1636 & 0.1643 & 0.2903 \\
\bottomrule
\end{tabular}
\end{table}

\begin{table}[H]
\centering
\caption{SVM (RBF Kernel) results across evaluation conditions.}
\label{tab:svm_results}
\small
\begin{tabular}{lrrrrr}
\toprule
\textbf{Condition} & \textbf{\auroc{}} & \textbf{Brier} & \textbf{Log Loss} & \textbf{Mean Human} & \textbf{Mean \llm{}} \\
\midrule
hc3\_to\_hc3   & 0.7993 & 0.1835 & 0.5486 & 0.3700 & 0.6318 \\
hc3\_to\_eli5  & 0.6933 & 0.2348 & 0.6639 & 0.5196 & 0.6686 \\
eli5\_to\_eli5 & 0.7924 & 0.1857 & 0.5512 & 0.3740 & 0.6287 \\
eli5\_to\_hc3  & 0.5992 & 0.3169 & 1.5852 & 0.2083 & 0.3191 \\
\bottomrule
\end{tabular}
\end{table}


\begin{figure}[H]
\centering
\includegraphics[width=0.9\textwidth]{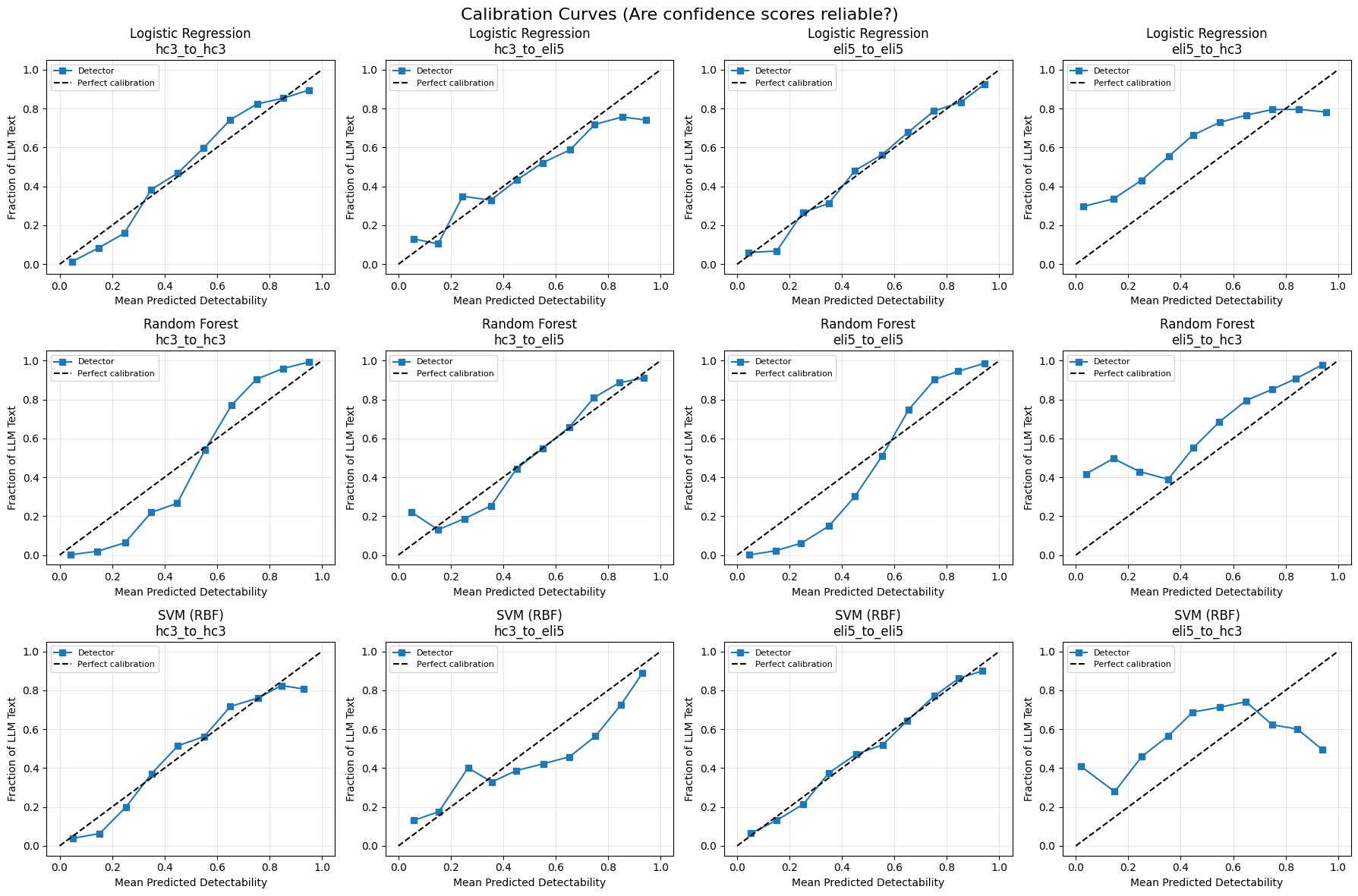}
\caption{Calibration curves for classical detectors across four evaluation settings.
Points close to the diagonal indicate well-calibrated confidence scores, while systematic
deviations reflect over- or under-confidence.}
\label{fig:classical_calibration}
\end{figure}

\paragraph{Key Observation.}
Random Forest achieves the strongest in-distribution performance (\auroc{} $= 0.977$ on
\hc{}) among classical detectors but suffers the largest cross-domain degradation
(eli5\_to\_hc3: $0.634$), suggesting it overfits to dataset-specific surface statistics
rather than generalizable linguistic signals.

\subsection{Fine-Tuned Encoder Transformers}

Tables~\ref{tab:bert_results}--\ref{tab:deberta_results} report full results for each
fine-tuned encoder.

\begin{table}[H]
\centering
\caption{BERT (\texttt{bert-base-uncased}) results.}
\label{tab:bert_results}
\small
\begin{tabular}{lrrrrrrr}
\toprule
\textbf{Condition} & \textbf{\auroc{}} & \textbf{Acc.} & \textbf{Brier} & \textbf{Log Loss} & \textbf{Hum.} & \textbf{\llm{}} & \textbf{Sep.} \\
\midrule
hc3\_to\_hc3   & 0.9947 & 0.9041 & 0.0906 & 0.5747 & 0.1927 & 0.9999 & 0.8071 \\
hc3\_to\_eli5  & 0.9489 & 0.8396 & 0.1472 & 0.8720 & 0.2319 & 0.9147 & 0.6828 \\
eli5\_to\_eli5 & 0.9943 & 0.9388 & 0.0572 & 0.3315 & 0.1245 & 0.9996 & 0.8751 \\
eli5\_to\_hc3  & 0.9083 & 0.8548 & 0.1393 & 0.8719 & 0.2100 & 0.9170 & 0.7070 \\
\bottomrule
\end{tabular}
\end{table}

\begin{table}[H]
\centering
\caption{RoBERTa (\texttt{roberta-base}) results.}
\label{tab:roberta_results}
\small
\begin{tabular}{lrrrrrrr}
\toprule
\textbf{Condition} & \textbf{\auroc{}} & \textbf{Acc.} & \textbf{Brier} & \textbf{Log Loss} & \textbf{Hum.} & \textbf{\llm{}} & \textbf{Sep.} \\
\midrule
hc3\_to\_hc3   & 0.9994 & 0.9679 & 0.0303 & 0.2204 & 0.0642 & 1.0000 & 0.9357 \\
hc3\_to\_eli5  & 0.9741 & 0.7967 & 0.1926 & 1.4401 & 0.4054 & 0.9991 & 0.5937 \\
eli5\_to\_eli5 & 0.9998 & 0.9645 & 0.0331 & 0.2264 & 0.0711 & 0.9999 & 0.9289 \\
eli5\_to\_hc3  & 0.9657 & 0.9045 & 0.0932 & 0.7082 & 0.1129 & 0.9214 & 0.8085 \\
\bottomrule
\end{tabular}
\end{table}

\begin{table}[H]
\centering
\caption{ELECTRA (\texttt{google/electra-base-discriminator}) results.}
\label{tab:electra_results}
\small
\begin{tabular}{lrrrrrrr}
\toprule
\textbf{Condition} & \textbf{\auroc{}} & \textbf{Acc.} & \textbf{Brier} & \textbf{Log Loss} & \textbf{Hum.} & \textbf{\llm{}} & \textbf{Sep.} \\
\midrule
hc3\_to\_hc3   & 0.9972 & 0.8639 & 0.1298 & 0.8663 & 0.2731 & 0.9996 & 0.7265 \\
hc3\_to\_eli5  & 0.9597 & 0.8492 & 0.1450 & 0.9868 & 0.2770 & 0.9725 & 0.6955 \\
eli5\_to\_eli5 & 0.9975 & 0.9605 & 0.0359 & 0.1804 & 0.0821 & 0.9986 & 0.9166 \\
eli5\_to\_hc3  & 0.9318 & 0.8790 & 0.1161 & 0.6408 & 0.1630 & 0.9140 & 0.7511 \\
\bottomrule
\end{tabular}
\end{table}

\begin{table}[H]
\centering
\caption{DistilBERT (\texttt{distilbert-base-uncased}) results.}
\label{tab:distilbert_results}
\small
\begin{tabular}{lrrrrrr}
\toprule
\textbf{Condition} & \textbf{\auroc{}} & \textbf{Acc.} & \textbf{Brier} & \textbf{Log Loss} & \textbf{Hum.} & \textbf{\llm{}} \\
\midrule
hc3\_to\_hc3   & 0.9968 & 0.9502 & 0.0460 & 0.2698 & 0.0999 & 0.9997 \\
hc3\_to\_eli5  & 0.9578 & 0.8835 & 0.1088 & 0.6235 & 0.1250 & 0.8907 \\
eli5\_to\_eli5 & 0.9983 & 0.9692 & 0.0288 & 0.1503 & 0.0647 & 0.9993 \\
eli5\_to\_hc3  & 0.9309 & 0.8702 & 0.1229 & 0.7205 & 0.1397 & 0.8768 \\
\bottomrule
\end{tabular}
\end{table}

\begin{table}[H]
\centering
\caption{DeBERTa-v3 (\texttt{microsoft/deberta-v3-base}) results.}
\label{tab:deberta_results}
\small
\begin{tabular}{lrrrrrr}
\toprule
\textbf{Condition} & \textbf{\auroc{}} & \textbf{Acc.} & \textbf{Brier} & \textbf{Log Loss} & \textbf{Hum.} & \textbf{\llm{}} \\
\midrule
hc3\_to\_hc3   & 0.9913 & 0.8888 & 0.1100 & 0.9803 & 0.2225 & 0.9991 \\
hc3\_to\_eli5  & 0.8762 & 0.5728 & 0.4245 & 4.0517 & 0.8532 & 0.9997 \\
eli5\_to\_eli5 & 0.9530 & 0.7794 & 0.2089 & 1.2387 & 0.4377 & 0.9998 \\
eli5\_to\_hc3  & 0.8890 & 0.7749 & 0.2148 & 1.3764 & 0.4147 & 0.9662 \\
\bottomrule
\end{tabular}
\end{table}

\begin{figure}[H]
\centering
\begin{subfigure}{\textwidth}
    \centering
    \includegraphics[width=0.9\textwidth]{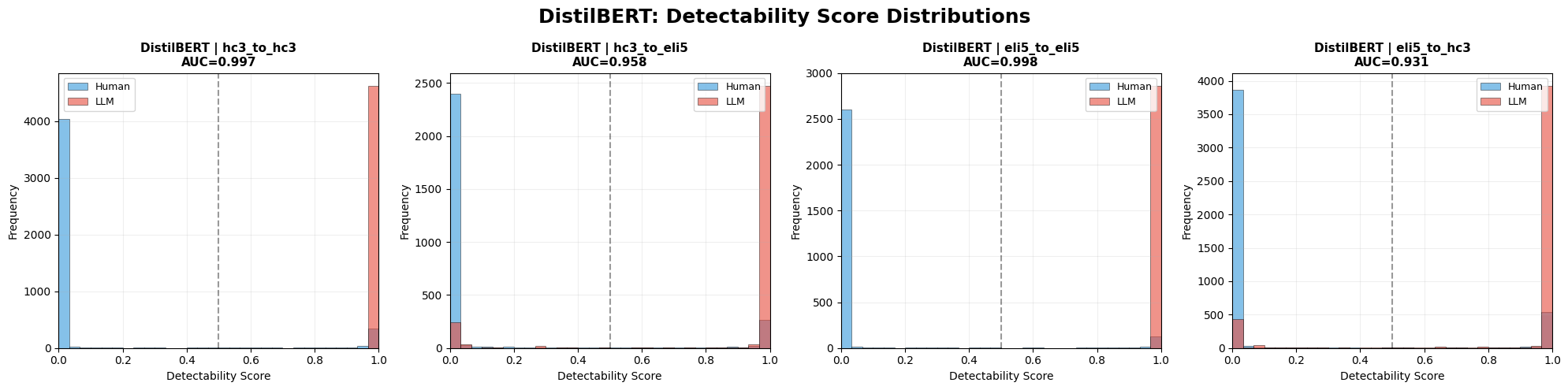}
    \caption{Detectability score distributions.}
\end{subfigure}
\vspace{0.5cm}
\begin{subfigure}{\textwidth}
    \centering
    \includegraphics[width=0.9\textwidth]{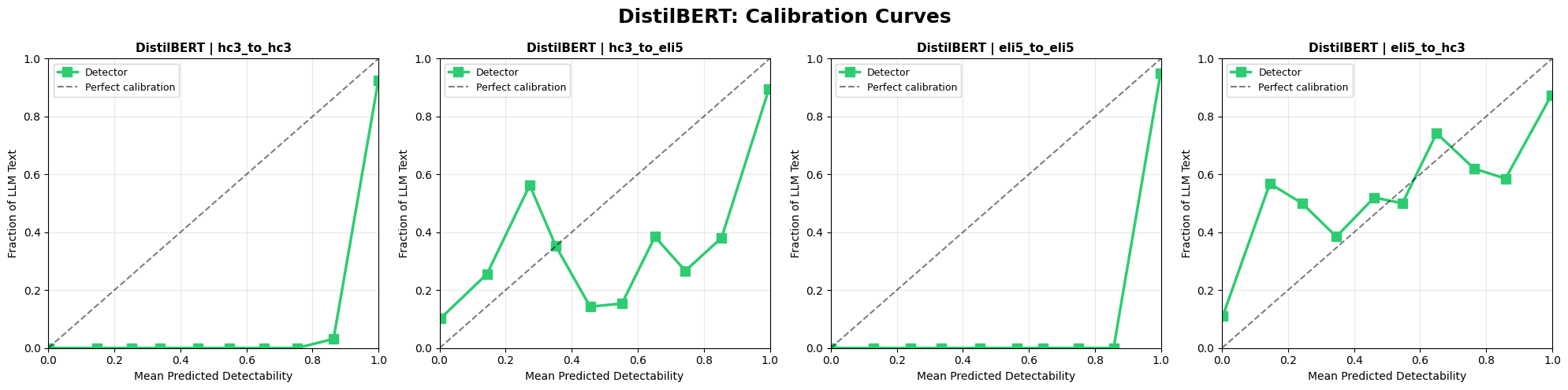}
    \caption{Calibration curves.}
\end{subfigure}
\vspace{0.5cm}
\begin{subfigure}{\textwidth}
    \centering
    \includegraphics[width=0.9\textwidth]{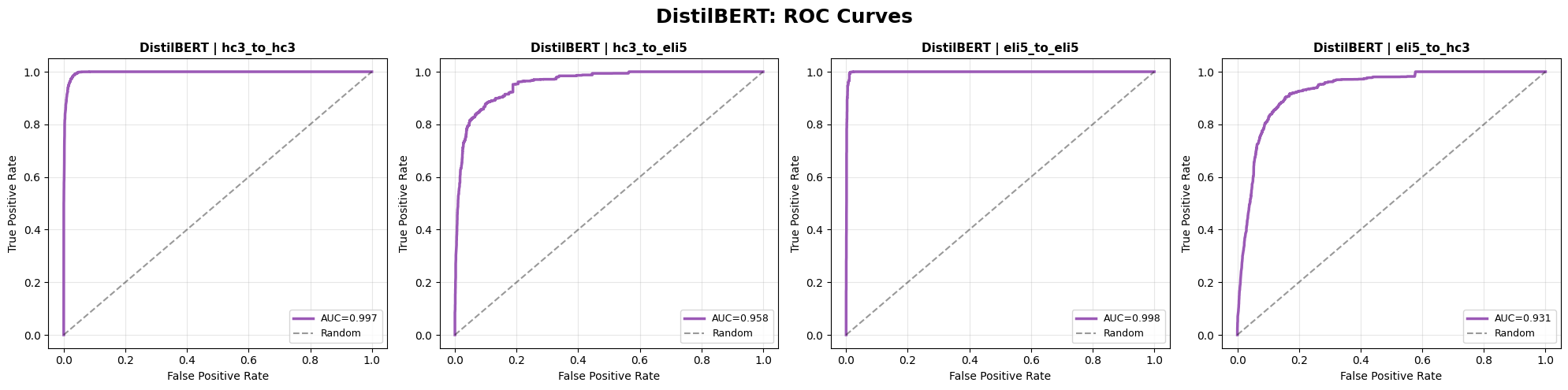}
    \caption{ROC curves.}
\end{subfigure}
\caption{Performance analysis of DistilBERT across four evaluation conditions.
Top: score distributions indicating class separability.
Middle: reliability diagrams assessing calibration.
Bottom: ROC curves illustrating discrimination performance.
DistilBERT achieves near-transformer performance at approximately 60\% of BERT's
parameter count.}
\label{fig:distilbert_results}
\end{figure}

\subsection{Shallow 1D-CNN Detector}

\begin{figure}[H]
\centering
\begin{subfigure}{\textwidth}
    \centering
    \includegraphics[width=0.9\textwidth]{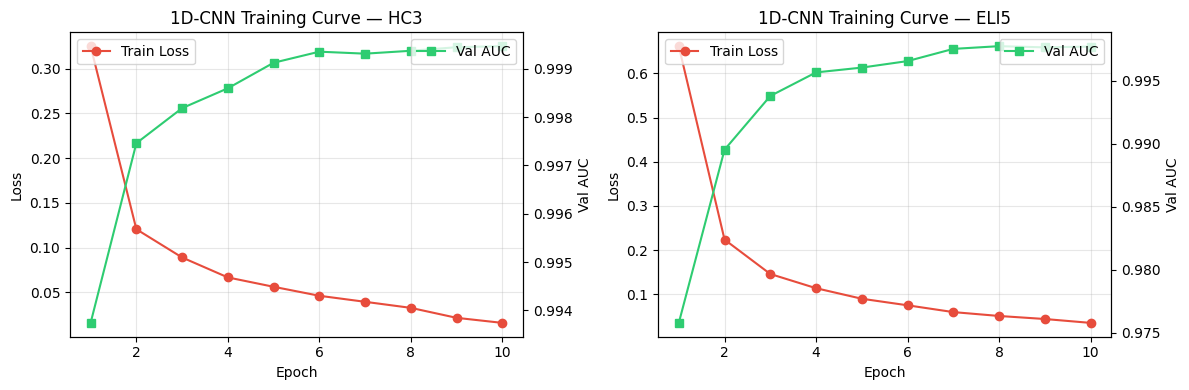}
    \caption{Training loss and validation AUC across epochs.}
\end{subfigure}
\vspace{0.5cm}
\begin{subfigure}{\textwidth}
    \centering
    \includegraphics[width=0.9\textwidth]{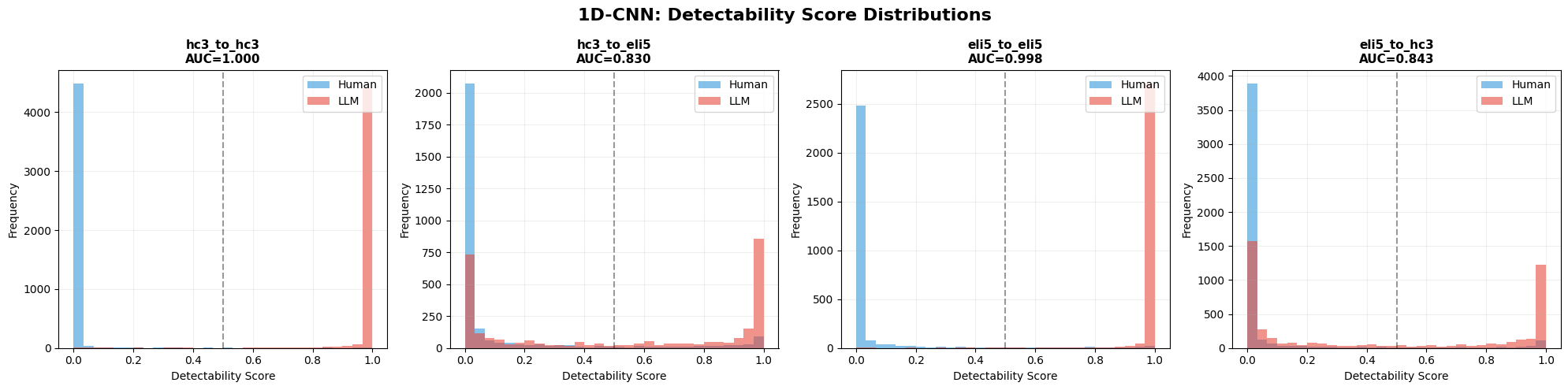}
    \caption{Detectability score distributions across evaluation conditions.}
\end{subfigure}
\caption{Training dynamics and detectability behavior of the 1D-CNN detector.
Top: rapid convergence to high validation AUC on both datasets.
Bottom: score distributions indicating strong separability between human and \llm{} text.}
\label{fig:cnn_training}
\end{figure}

\begin{figure}[H]
\centering
\includegraphics[width=0.7\textwidth]{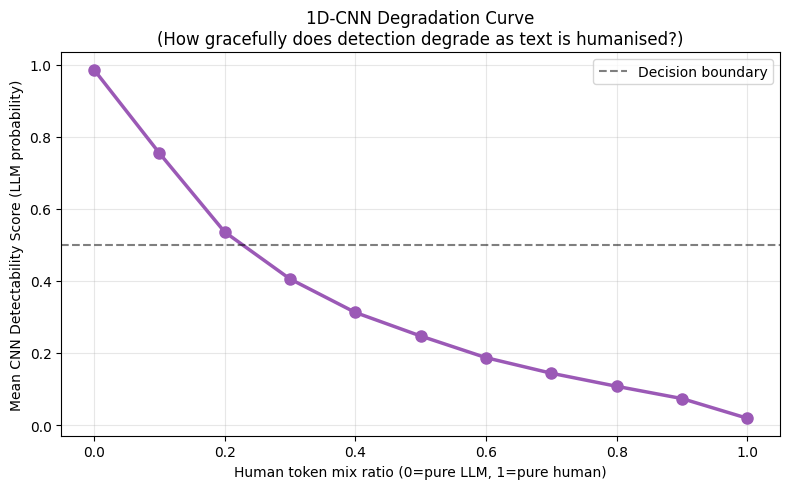}
\caption{1D-CNN degradation curve under progressive text humanization.  The $x$-axis
represents the fraction of human tokens mixed into otherwise \llm{}-generated text.
The steep, smooth decline confirms that the 1D-CNN is highly sensitive to even small
amounts of human-style $n$-gram patterns.}
\label{fig:cnn_degradation}
\end{figure}

Table~\ref{tab:cnn_results} reports 1D-CNN results.
Figure~\ref{fig:cnn_training} shows training curves and score distributions;
Figure~\ref{fig:cnn_degradation} shows the degradation curve under progressive humanization.

\begin{table}[H]
\centering
\caption{1D-CNN results across evaluation conditions.}
\label{tab:cnn_results}
\small
\begin{tabular}{lrrrrrr}
\toprule
\textbf{Condition} & \textbf{\auroc{}} & \textbf{Acc.} & \textbf{Brier} & \textbf{Log Loss} & \textbf{Hum.} & \textbf{\llm{}} \\
\midrule
hc3\_to\_hc3   & 0.9995 & 0.9916 & 0.0067 & 0.0262 & 0.0093 & 0.9862 \\
hc3\_to\_eli5  & 0.8303 & 0.7124 & 0.2446 & 1.0887 & 0.1192 & 0.5275 \\
eli5\_to\_eli5 & 0.9982 & 0.9748 & 0.0191 & 0.0666 & 0.0477 & 0.9844 \\
eli5\_to\_hc3  & 0.8432 & 0.6866 & 0.2752 & 1.4723 & 0.0730 & 0.4455 \\
\bottomrule
\end{tabular}
\end{table}

\paragraph{Key Observation.}
The 1D-CNN achieves near-perfect in-distribution \auroc{} ($0.9995$ on \hc{}) ---
competitive with full transformers --- while having $20\times$ fewer parameters.
Cross-domain performance drops to $0.83$--$0.84$, indicating that learned $n$-gram
patterns are domain-specific but still substantially more transferable than pure
classical features.

\subsection{Stylometric and Statistical Hybrid Detector}

\begin{figure}[H]
\centering
\includegraphics[width=0.7\textwidth]{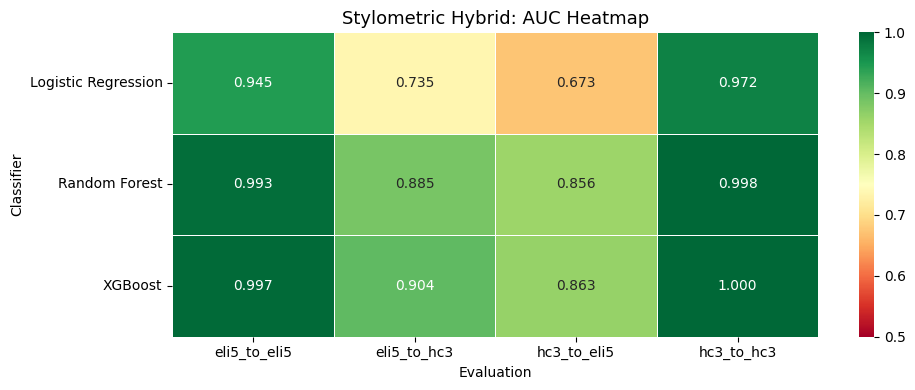}
\caption{Stylometric hybrid \auroc{} heatmap.  Rows correspond to classifiers
(Logistic Regression, Random Forest, XGBoost), while columns represent the four
evaluation conditions (eli5-to-eli5, eli5-to-hc3, hc3-to-eli5, hc3-to-hc3).
Cell colors range from red (0.5) to dark green (1.0).  XGBoost dominates across
all conditions; the cross-domain eli5-to-hc3 \auroc{} of $0.904$ represents a
substantial improvement over the classical Stage~1 Random Forest ($0.634$).}
\label{fig:stylo_heatmap}
\end{figure}

Tables~\ref{tab:stylo_lr}--\ref{tab:stylo_xgb} report results for all three classifiers
trained on the extended stylometric feature set.  Figure~\ref{fig:stylo_heatmap} shows
the \auroc{} heatmap across classifiers and evaluation conditions, and

\begin{table}[H]
\centering
\caption{Stylometric Hybrid --- Logistic Regression results.}
\label{tab:stylo_lr}
\small
\begin{tabular}{lrrrrrrrr}
\toprule
\textbf{Condition} & \textbf{\auroc{}} & \textbf{Acc.} & \textbf{Brier} & \textbf{Log Loss} & \textbf{Hum.} & \textbf{\llm{}} & \textbf{Sep.} \\
\midrule
hc3\_to\_hc3   & 0.9721 & 0.9243 & 0.0580 & 0.2093 & 0.1273 & 0.8753 & 0.7480 \\
hc3\_to\_eli5  & 0.6731 & 0.6296 & 0.2539 & 0.8110 & 0.3668 & 0.5502 & 0.1834 \\
eli5\_to\_eli5 & 0.9448 & 0.8807 & 0.0897 & 0.3003 & 0.1823 & 0.8166 & 0.6343 \\
eli5\_to\_hc3  & 0.7348 & 0.6650 & 0.2669 & 1.2185 & 0.2006 & 0.4941 & 0.2935 \\
\bottomrule
\end{tabular}
\end{table}

\begin{table}[H]
\centering
\caption{Stylometric Hybrid --- Random Forest results.}
\label{tab:stylo_rf}
\small
\begin{tabular}{lrrrrrrr}
\toprule
\textbf{Condition} & \textbf{\auroc{}} & \textbf{Acc.} & \textbf{Brier} & \textbf{Log Loss} & \textbf{Hum.} & \textbf{\llm{}} & \textbf{Sep.} \\
\midrule
hc3\_to\_hc3   & 0.9981 & 0.9785 & 0.0189 & 0.0854 & 0.0731 & 0.9362 & 0.8631 \\
hc3\_to\_eli5  & 0.8557 & 0.7516 & 0.1768 & 0.5586 & 0.1699 & 0.5628 & 0.3929 \\
eli5\_to\_eli5 & 0.9934 & 0.9605 & 0.0395 & 0.1626 & 0.1371 & 0.8759 & 0.7388 \\
eli5\_to\_hc3  & 0.8848 & 0.6589 & 0.2100 & 0.6123 & 0.1164 & 0.4363 & 0.3199 \\
\bottomrule
\end{tabular}
\end{table}

\begin{table}[H]
\centering
\caption{Stylometric Hybrid --- XGBoost results.}
\label{tab:stylo_xgb}
\small
\begin{tabular}{lrrrrrrr}
\toprule
\textbf{Condition} & \textbf{\auroc{}} & \textbf{Acc.} & \textbf{Brier} & \textbf{Log Loss} & \textbf{Hum.} & \textbf{\llm{}} & \textbf{Sep.} \\
\midrule
hc3\_to\_hc3   & 0.9996 & 0.9928 & 0.0059 & 0.0226 & 0.0179 & 0.9912 & 0.9733 \\
hc3\_to\_eli5  & 0.8633 & 0.7252 & 0.2270 & 0.9451 & 0.0673 & 0.5033 & 0.4361 \\
eli5\_to\_eli5 & 0.9971 & 0.9732 & 0.0197 & 0.0714 & 0.0529 & 0.9620 & 0.9091 \\
eli5\_to\_hc3  & 0.9037 & 0.7275 & 0.2281 & 0.9624 & 0.0439 & 0.4808 & 0.4369 \\
\bottomrule
\end{tabular}
\end{table}

\paragraph{Key Observation.}
XGBoost on the full stylometric feature set achieves \auroc{} $= 0.9996$ in-distribution
--- on par with fine-tuned transformers --- while remaining fully interpretable.
The extended feature set (particularly sentence-level perplexity CV, connector density,
and AI-phrase density) substantially improves cross-domain performance over the classical
Stage~1 feature set alone, with XGBoost eli5\_to\_hc3 reaching $0.904$ versus Random
Forest's $0.634$ in the classical setting.

\subsection{Stage 1 Key Conclusions}
\label{sec:stage1_conclusions}

\begin{enumerate}[leftmargin=*,label=\textbf{\arabic*.}]
  \item \textbf{Fine-tuned encoder transformers dominate all other families.}  RoBERTa
        achieves the highest in-distribution \auroc{} ($0.9994$ on \hc{}), confirming that
        task-specific fine-tuning on paired human/\llm{} data is the most effective
        detection strategy.
  \item \textbf{Cross-domain degradation is universal and substantial.}  Every detector
        family suffers \auroc{} drops of 5--30 points when trained on one dataset and tested
        on the other, indicating that no current detector generalizes robustly across \llm{}
        sources and domains.
  \item \textbf{The 1D-CNN achieves near-transformer in-distribution performance with
        $20\times$ fewer parameters.}  Its cross-domain performance ($0.83$--$0.84$)
        reveals that learned $n$-gram patterns are dataset-specific rather than universally
        generalizable.
  \item \textbf{DeBERTa-v3 is competitive in-distribution but severely miscalibrated
        cross-domain.}  Following FP32 precision correction, it reaches \auroc{} $0.991$
        (\hc{}) and $0.953$ (\eli{}) in-distribution.  Cross-domain transfer exposes a
        critical failure: hc3\_to\_eli5 accuracy collapses to $0.573$ (log loss $4.052$)
        despite \auroc{} $0.876$, indicating well-ordered but poorly calibrated scores ---
        consistent with overfitting to \hc{}'s formal register.
  \item \textbf{The XGBoost stylometric hybrid matches transformer in-distribution
        performance while remaining fully interpretable.}  Sentence-level perplexity CV,
        connector density, and AI-phrase density are the most discriminative features.
  \item \textbf{Length matching was critical.}  Without the $\pm20\%$ length normalization,
        classical detectors would have trivially exploited the well-known length disparity
        between human and \llm{} answers, inflating reported performance.
\end{enumerate}

\section{LLM-as-Detector and Contrastive Likelihood Detection}
\label{sec:llm_detector}

This section evaluates generative \llm{}s as zero-parameter AI-text detectors across six
model scales --- from sub-2B to a frontier API model --- under three prompting regimes.
The pipeline incorporates calibrated threshold analysis alongside fixed-threshold
evaluation, and a hybrid confidence-logit scoring scheme for Chain-of-Thought outputs.

\subsection{Prompting Paradigms}
\label{subsec:prompting_paradigms}

\textbf{Zero-Shot} prompting presents only a system instruction and target text.
Detection scores are derived via constrained decoding: the next-token log-probability
distribution is read at the final prompt position, and a soft $[0,1]$ detectability score
is computed from the softmax of $\log P(\texttt{llm})$ versus $\log P(\texttt{human})$,
yielding a continuous score without generation.

\textbf{Few-Shot} prompting augments the zero-shot prompt with $k$ labeled examples
drawn from the training pool ($k=3$ for sub-2B models; $k=5$ otherwise), with TF-IDF
based semantic retrieval used for larger models to select maximally informative
demonstrations.

\textbf{Chain-of-Thought (\CoT{})} prompting instructs the model to reason across
structured linguistic dimensions before delivering a final \texttt{VERDICT}.  \CoT{}
scoring employs a hybrid confidence-logit scheme: when the model produces a parseable
numerical confidence estimate alongside its verdict, this is combined with the
logit-derived soft score at the verdict token position using equal weighting; otherwise,
the logit-only score is used.  \CoT{} is restricted to models with sufficient
instruction-following capacity; sub-2B models are excluded.  Three threshold strategies
are reported: fixed at 0.5 (\texttt{acc@0.5}), calibrated at the score median
(\texttt{acc@median}), and optimal Youden-J (\texttt{acc@optimal}).

\subsection{Tiny-Scale Models: TinyLlama-1.1B-Chat-v1.0and Qwen2.5-1.5B}
\label{subsec:tiny_models}

Both models are evaluated under zero-shot and few-shot regimes on 500 balanced samples
per dataset, loaded in FP16 with \texttt{device\_map="auto"}.

\begin{table}[ht]
\centering
\caption{Tiny-scale \llm{}-as-detector results.}
\label{tab:tiny_detector}
\resizebox{\textwidth}{!}{%
\begin{tabular}{llccccc}
\toprule
\textbf{Model} & \textbf{Regime} & \textbf{Dataset} & \textbf{\auroc{}} &
\textbf{acc@0.5} & \textbf{acc@median} & \textbf{acc@optimal} \\
\midrule
TinyLlama-1.1B-Chat-v1.0& Zero-Shot & HC3    & 0.5653 & 0.558 & 0.534 & 0.558 \\
TinyLlama-1.1B-Chat-v1.0& Zero-Shot & \eli{} & 0.5072 & 0.524 & 0.510 & 0.524 \\
TinyLlama-1.1B-Chat-v1.0& Few-Shot  & HC3    & 0.6198 & 0.614 & 0.600 & 0.614 \\
TinyLlama-1.1B-Chat-v1.0& Few-Shot  & \eli{} & 0.5860 & 0.580 & 0.566 & 0.580 \\
\midrule
Qwen2.5-1.5B-Instruct  & Zero-Shot & HC3    & 0.5221 & 0.436 & 0.530 & 0.574 \\
Qwen2.5-1.5B-Instruct  & Zero-Shot & \eli{} & 0.5205 & 0.470 & 0.512 & 0.562 \\
Qwen2.5-1.5B-Instruct  & Few-Shot  & HC3    & 0.4794 & 0.450 & 0.518 & 0.536 \\
Qwen2.5-1.5B-Instruct  & Few-Shot  & \eli{} & 0.6340 & 0.484 & 0.620 & 0.620 \\
\bottomrule
\end{tabular}}
\end{table}

Both models perform near chance across all conditions (\auroc{} 0.48--0.63), confirming
that detection as a meta-cognitive task does not emerge at sub-2B scale.  The threshold
analysis surfaces a qualitatively important finding: Qwen2.5-1.5B-Instructzero-shot scores cluster
systematically above 0.5 (median $\approx$ 0.75--0.80), yet \auroc{} remains near chance
--- a \textit{score-collapsing} pattern in which the model emits uniformly high
detectability scores regardless of label, yielding poor rank ordering rather than a polarity
inversion.  TinyLlama's few-shot median score shifts from $\approx\!0.26$ (zero-shot) to
$\approx\!0.69$ (few-shot), reflecting a format-induced distributional shift rather than
improved class discrimination.

\subsection{Mid-Scale Models:Llama-3.1-8B-Instruct and Qwen2.5-7B}
\label{subsec:mid_models}

Both 8B models are evaluated under all three regimes.  Zero-shot and few-shot use
constrained decoding on 500 samples; \CoT{} uses full autoregressive generation
(max 400 tokens, greedy decoding) on 70 samples.

\begin{table}[ht]
\centering
\caption{Mid-scale \llm{}-as-detector results.  FP/FN denote false positive/negative counts.}
\label{tab:mid_detector}
\resizebox{\textwidth}{!}{%
\begin{tabular}{llcccccc}
\toprule
\textbf{Model} & \textbf{Regime} & \textbf{Dataset} & \textbf{\auroc{}} &
\textbf{acc@0.5} & \textbf{acc@optimal} & \textbf{FP} & \textbf{FN} \\
\midrule
Llama-3.1-8B-Instruct& Zero-Shot & HC3    & 0.7295 & 0.680 & 0.680 & 48  & 120 \\
Llama-3.1-8B-Instruct& Zero-Shot & \eli{} & 0.7508 & 0.670 & 0.702 & 106 & 57  \\
Llama-3.1-8B-Instruct& Few-Shot  & HC3    & 0.5027 & 0.546 & 0.550 & 53  & 172 \\
Llama-3.1-8B-Instruct& Few-Shot  & \eli{} & 0.5961 & 0.574 & 0.578 & 77  & 140 \\
Llama-3.1-8B-Instruct& \CoT{}    & HC3    & 0.6771 & 0.629 & 0.657 & 11  & 18  \\
Llama-3.1-8B-Instruct& \CoT{}    & \eli{} & 0.5988 & 0.586 & 0.600 & 15  & 13  \\
\midrule
Qwen2.5-7B-Instruct  & Zero-Shot & HC3    & 0.6902 & 0.656 & 0.666 & 70  & 102 \\
Qwen2.5-7B-Instruct  & Zero-Shot & \eli{} & 0.6638 & 0.620 & 0.632 & 60  & 130 \\
Qwen2.5-7B-Instruct  & Few-Shot  & HC3    & 0.4579 & 0.484 & 0.502 & 132 & 126 \\
Qwen2.5-7B-Instruct  & Few-Shot  & \eli{} & 0.5042 & 0.524 & 0.542 & 125 & 113 \\
Qwen2.5-7B-Instruct  & \CoT{}    & HC3    & 0.6388 & 0.514 & 0.657 & 31  & 2   \\
Qwen2.5-7B-Instruct  & \CoT{}    & \eli{} & 0.7808 & 0.614 & 0.743 & 20  & 3   \\
\bottomrule
\end{tabular}}
\end{table}

Llama-3.1-8B-Instructachieves competitive zero-shot \auroc{} of 0.730--0.751, demonstrating that
genuine detection signal emerges at 8B scale without in-context examples.  However,
few-shot prompting markedly degrades performance (\auroc{} 0.503--0.596).  Qwen2.5-7B
\CoT{} on \eli{} achieves 0.781, the highest among mid-scale models.

\subsection{Large-Scale Models: LLaMA-2-13B-Chat}
\label{subsec:llama2_13b}

\subsubsection*{Pipeline Design}

Llama-2-13b-chat-hf is evaluated under all three regimes, loaded with 4-bit NF4 quantization
(double quantization, FP16 compute dtype).  Zero-shot and few-shot use 200 samples per
dataset; \CoT{} uses 30 samples with full generation (max 400 tokens, greedy decoding).

Token-level debug analysis revealed that Llama-2-13b-chat-hfexhibits a strong unconditional
``no'' bias.  Rather than inverting this post-hoc, the pipeline resolves it structurally
via \textbf{prompt polarity swapping}: the model is asked ``Was this text written by a
human?'' with $\texttt{yes} = \text{human}$ and $\texttt{no} = \text{AI-generated}$.
Additionally, a \textbf{task-specific prior} is computed by averaging yes/no logits over
50 real task prompts drawn from the evaluation pool; subtracting this prior removes
task-level marginal bias while preserving sample-discriminative signal.

The \CoT{} prompt frames the task as a \textit{stylometric analysis} --- avoiding the
term ``AI detection'' to circumvent LLaMA-2's safety-oriented refusal behaviors.  The
model scores seven linguistic dimensions on a 0--10 scale.  The hybrid confidence-logit
ensemble weights confirmed confidence and logit scores at 0.6/0.4 when confidence falls
outside the dead zone $[0.40, 0.60]$; otherwise the logit score is used alone.

\begin{table}[ht]
\centering
\caption{Llama-2-13b-chat-hfdetection results ($n=200$ zero/few-shot; $n=30$ \CoT{}).}
\label{tab:llama2_results}
\resizebox{\textwidth}{!}{%
\begin{tabular}{llcccccc}
\toprule
\textbf{Regime} & \textbf{Dataset} & \textbf{\auroc{}} & \textbf{acc@0.5} &
\textbf{acc@median} & \textbf{acc@optimal} & \textbf{FP} & \textbf{FN} \\
\midrule
Zero-Shot & HC3    & 0.8124 & 0.715 & 0.710 & 0.760 & 21 & 36 \\
Zero-Shot & \eli{} & 0.8098 & 0.750 & 0.755 & 0.760 & 28 & 22 \\
Few-Shot  & HC3    & 0.6678 & 0.635 & 0.630 & 0.660 & 28 & 45 \\
Few-Shot  & \eli{} & 0.6374 & 0.590 & 0.620 & 0.620 & 38 & 44 \\
\CoT{}    & HC3    & 0.8778 & 0.833 & 0.867 & 0.867 &  2 &  4 \\
\CoT{}    & \eli{} & 0.8978 & 0.733 & 0.800 & 0.867 &  2 &  3 \\
\bottomrule
\end{tabular}}
\end{table}

The corrected pipeline yields substantially improved results relative to the original
implementation (\auroc{} 0.363--0.705 attributable to polarity and prior
misconfiguration).  Zero-shot \auroc{} of 0.810--0.812 is consistent across both datasets,
and \CoT{} peaks at 0.878 and 0.898 on \hc{} and \eli{} respectively --- the strongest
\CoT{} results among all open-source models.

\subsection{Large-Scale Models: Qwen2.5-14B-Instruct}
\label{subsec:qwen25_14b}

\subsubsection*{Pipeline Design}

Qwen2.5-14B-Instruct is evaluated under all three regimes using the same swapped polarity
convention, loaded with 4-bit NF4 quantization and BFloat16 compute dtype.  The original
implementation suffered from 86.7--90.0\% unknown rates due to two failure modes: premature
generation termination when \texttt{eos\_token\_id} was used as \texttt{pad\_token\_id},
and an insufficient \texttt{max\_new\_tokens=350} budget.  The corrected pipeline sets
\texttt{pad\_token\_id} explicitly and increases \texttt{max\_new\_tokens} to 500.
Task framing adopts ``forensic linguist performing authorship attribution analysis'' to
minimize safety-motivated refusals.

\begin{table}[ht]
\centering
\caption{Qwen2.5-14B-Instruct detection results ($n=200$ zero/few-shot; $n=30$ \CoT{}).}
\label{tab:qwen14b_results}
\resizebox{\textwidth}{!}{%
\begin{tabular}{llcccccc}
\toprule
\textbf{Regime} & \textbf{Dataset} & \textbf{\auroc{}} & \textbf{acc@0.5} &
\textbf{acc@median} & \textbf{acc@optimal} & \textbf{FP} & \textbf{FN} \\
\midrule
Zero-Shot & HC3    & 0.6686 & 0.680 & 0.660 & 0.680 & 31 & 33 \\
Zero-Shot & \eli{} & 0.7294 & 0.690 & 0.655 & 0.695 & 41 & 21 \\
Few-Shot  & HC3    & 0.3153 & 0.385 & 0.390 & 0.500 & 52 & 71 \\
Few-Shot  & \eli{} & 0.4262 & 0.470 & 0.465 & 0.500 & 59 & 47 \\
\CoT{}    & HC3    & 0.6622 & 0.700 & 0.667 & 0.733 &  4 &  4 \\
\CoT{}    & \eli{} & 0.8000 & 0.733 & 0.667 & 0.767 &  4 &  3 \\
\bottomrule
\end{tabular}}
\end{table}

\begin{table}[ht]
\centering
\caption{Qwen2.5-14B-Instruct\CoT{} component analysis: hybrid vs.\ logit-only scoring.}
\label{tab:qwen14b_cot_component}
\begin{tabular}{llcccc}
\toprule
\textbf{Dataset} & \textbf{Score Type} & \textbf{$n$} & \textbf{\auroc{}} &
\textbf{Accuracy} & \textbf{\% of Total} \\
\midrule
HC3    & conf+logit  & 11 & 0.6250 & 0.727 & 36.7\% \\
HC3    & logit\_only & 19 & 0.6429 & 0.684 & 63.3\% \\
\eli{} & conf+logit  & 15 & 0.8519 & 0.800 & 50.0\% \\
\eli{} & logit\_only & 15 & 0.7593 & 0.667 & 50.0\% \\
\bottomrule
\end{tabular}
\end{table}

\subsection{GPT-4o-mini as Detector}
\label{subsec:gpt4omini}

GPT-4o-mini is evaluated via the OpenAI API using a structured 7-dimension rubric scoring
protocol across three regimes.  Unlike local models --- where constrained logit decoding
is used --- GPT-4o-mini employs a \textbf{rubric-based elicitation strategy} that forces
the model to commit to seven independent dimension scores (hedging/formulaic language,
response completeness, personal voice, lexical uniformity, structural neatness, response
fit, formulaic tells) before producing a final \texttt{AI\_SCORE} $\in [0, 100]$.  This
design circumvents the known \textsc{rlhf}-induced suppression of numeric probability
outputs.  A full five-metric evaluation suite is applied with 1{,}000-iteration bootstrap
confidence intervals ($n = 200$ for zero-shot/few-shot; $n = 50$ for \CoT{}).

\begin{table}[H]
\centering
\caption{GPT-4o-mini (\llm{}-as-detector) results.  All score directions verified correct.}
\label{tab:gpt4omini_results}
\small
\begin{tabular}{lllrrrr}
\toprule
\textbf{Model} & \textbf{Regime} & \textbf{Data} & \textbf{\auroc{}} &
\textbf{Acc.@0.5} & \textbf{Acc.@Opt.} & \textbf{Sep.} \\
\midrule
GPT-4o-mini & ZS     & \hc{}  & 0.8470 & 0.7600 & 0.7900 & $+$0.311 \\
GPT-4o-mini & ZS     & \eli{} & 0.9093 & 0.8800 & 0.8800 & $+$0.419 \\
GPT-4o-mini & FS     & \hc{}  & 0.7163 & 0.7000 & 0.7000 & $+$0.187 \\
GPT-4o-mini & FS     & \eli{} & 0.7824 & 0.6800 & 0.7400 & $+$0.246 \\
GPT-4o-mini & \CoT{} & \hc{}  & 0.8056 & 0.7800 & 0.8000 & $+$0.279 \\
GPT-4o-mini & \CoT{} & \eli{} & 0.7744 & 0.7600 & 0.8000 & $+$0.261 \\
\bottomrule
\end{tabular}
\end{table}

\paragraph{Finding 1: Structured rubric prompting outperforms constrained decoding.}
GPT-4o-mini achieves the highest zero-shot \auroc{} of all five models evaluated
($0.8470$ vs.\ $0.8124$ on \hc{}; $0.9093$ vs.\ $0.8098$ on \eli{} relative to
LLaMA-2-13B).

\paragraph{Finding 2: GPT-4o-mini degrades least under few-shot prompting.}

\begin{table}[H]
\centering
\caption{Zero-shot to few-shot \auroc{} degradation on \hc{}.}
\label{tab:fs_degradation}
\small
\begin{tabular}{lrrr}
\toprule
\textbf{Model} & \textbf{ZS \auroc{}} & \textbf{FS \auroc{}} & $\Delta$ \\
\midrule
Qwen2.5-14B-Instruct         & 0.6686 & 0.3153 & $-0.353$ \\
Qwen2.5-7B-Instruct          & 0.6902 & 0.4579 & $-0.232$ \\
Llama-3.1-8B-Instruct        & 0.7295 & 0.5027 & $-0.227$ \\
Llama-2-13b-chat-hf          & 0.8124 & 0.6678 & $-0.145$ \\
\textbf{GPT-4o-mini} & \textbf{0.8470} & \textbf{0.7163} & $\mathbf{-0.131}$ \\
\bottomrule
\end{tabular}
\end{table}

\paragraph{Finding 3: \CoT{} underperforms zero-shot for GPT-4o-mini.}
\auroc{} drops from $0.8470$ to $0.8056$ on \hc{} ($\Delta = -0.041$) and from
$0.9093$ to $0.7744$ on \eli{} ($\Delta = -0.135$).  Adding per-dimension reasoning
to an already-explicit rubric introduces noise rather than precision.

\paragraph{Finding 4: \eli{} is easier than \hc{} under zero-shot.}
GPT-4o-mini achieves \auroc{} $0.9093$ on \eli{} versus $0.8470$ on \hc{}
($\Delta = +0.062$).  \eli{}'s Mistral-7B-generated text carries stronger stylistic
markers than \hc{}'s ChatGPT-3.5 text.

\subsection*{Stage 1b Conclusions}
\label{subsec:stage1b_conclusions}

\textbf{Detection capability scales non-monotonically with parameter count.}  Sub-2B
models perform near random (\auroc{} 0.48--0.63); meaningful discrimination first appears
at 8B and consolidates at 13B (Llama-2-13b-chat-hf zero-shot: 0.810--0.812).  Qwen2.5-14B
zero-shot (0.669--0.729) underperforms Llama-2-13b-chat-hf at the same regime, indicating that
RLHF alignment strategy and prompt polarity interact with scale in ways that confound
simple parameter-count comparisons.

\textbf{Prompt polarity correction and task prior subtraction are necessary conditions
for valid constrained decoding.}  Naive constrained decoding without prior correction
produces systematically inverted or near-random scores due to RLHF-induced unconditional
response biases.

\textbf{\CoT{} prompting provides the largest and most consistent gains, contingent on
correct implementation.}  Llama-2-13b-chat-hf \CoT{} peaks at \auroc{} 0.878--0.898,
Qwen2.5-7B-Instruct\CoT{} reaches 0.781 on \eli{}.  \CoT{} gains are contingent on sufficient
generation budget, correct \texttt{pad\_token\_id} handling, safety-neutral prompt framing,
and a robust multi-fallback verdict parser.

\textbf{Few-shot prompting is consistently harmful across all model scales.}

Few-shot degrades \auroc{} relative to zero-shot:Llama-3.1-8B-Instruct (0.503--0.596 vs.
0.730--0.751), Llama-2-13b-chat-hf (0.637--0.668 vs.\ 0.810--0.812), Qwen2.5-14B-Instruct(0.315--0.426),
and GPT-4o-mini on \hc{} (0.7163 vs.\ 0.8470).

\textbf{The generator--detector identity confound is critical.}  Mistral-7B-Instruct,
used to generate the \eli{} \llm{} answers, performed near or below random as a detector
(\auroc{} 0.363--0.540).  A model cannot reliably detect its own outputs.

\textbf{No \llm{}-as-detector configuration approaches supervised fine-tuned encoders.}
The best result --- GPT-4o-mini zero-shot on \eli{} at \auroc{} $0.9093$ --- remains
well below RoBERTa in-distribution (\auroc{} $0.9994$).

\subsection{Contrastive Likelihood Detection}
\label{subsec:contrastive}

The contrastive score is defined as:
\begin{equation}
S(x) = \log P_{\text{large}}(x) - \log P_{\text{small}}(x)
\end{equation}

\begin{table}[H]
\centering
\caption{Contrastive likelihood detection results.}
\label{tab:contrastive_results}
\small
\begin{tabular}{llcc}
\toprule
\textbf{Variant} & \textbf{Dataset} & \textbf{\auroc{}} & \textbf{Score Sep.} \\
\midrule
\texttt{base\_contrast}  & HC3    & 0.5007 & 0.0007 \\
\texttt{base\_contrast}  & \eli{} & 0.6873 & 0.1873 \\
\texttt{multi\_scale}    & HC3    & 0.5007 & 0.0007 \\
\texttt{multi\_scale}    & \eli{} & 0.6873 & 0.1873 \\
\texttt{token\_variance} & HC3    & 0.6323 & 0.1323 \\
\texttt{token\_variance} & \eli{} & 0.5644 & 0.0644 \\
\textbf{\texttt{hybrid}} & HC3    & 0.5999 & 0.0446 \\
\textbf{\texttt{hybrid}} & \eli{} & \textbf{0.7615} & \textbf{0.1463} \\
\bottomrule
\end{tabular}
\end{table}

The hybrid score achieves \auroc{} of 0.762 on \eli{} but near-random on \hc{} (0.600
and below).  The performance gap is explained by a \textit{representational affinity
constraint}: GPT-2 and Mistral-7B share architectural and pretraining characteristics,
while ChatGPT (GPT-3.5) underwent extensive RLHF alignment at a larger parameter scale.

\section{Perplexity-Based Detectors}
\label{sec:perplexity}

\subsection{Method}

Perplexity-based detection is unsupervised and training-free.  Because GPT-2 and GPT-Neo
family models assign systematically \emph{lower} perplexity to \llm{}-generated text than
to human-written text, the detectability score is an \textbf{inversion} of the raw
perplexity signal:
\begin{equation}
\text{PPL}(x) = \exp\!\left(-\frac{1}{T}\sum_{t=1}^{T}
\log P(x_t \mid x_1, \ldots, x_{t-1})\right)
\end{equation}

Five reference models are evaluated: GPT-2 Small (117M), GPT-2 Medium (345M),
GPT-2 XL (1.5B), GPT-Neo-125M, and GPT-Neo-1.3B.  All models are run in FP16 on the
full \hc{} and \eli{} test sets.  A sliding window (512-token window, 256-token stride)
handles long texts.  Outlier perplexities are clipped at 10{,}000 for rank stability.

Raw perplexities are converted to $[0,1]$ detectability scores via four normalization
methods: \textit{rank-based}, \textit{log-rank}, \textit{minmax}, and \textit{sigmoid}.
The best method per condition is selected by \auroc{}, with optimal decision thresholds
identified via Youden's~$J$ statistic.

\subsection{Results}

\begin{table}[H]
\centering
\caption{Perplexity-based detector results.  Method = best normalization by \auroc{}.}
\label{tab:ppl_results}
\small
\begin{tabular}{llllrrrrr}
\toprule
\textbf{Model} & \textbf{Data} & \textbf{Method} &
\textbf{\auroc{}} & \textbf{Brier} & \textbf{Acc@Opt} &
\textbf{Hum.} & \textbf{\llm{}} & \textbf{Sep.} \\
\midrule
GPT-2 Small  & \hc{}  & rank   & 0.9099 & 0.1284 & 0.8805 & 0.2950 & 0.7050 & 0.4100 \\
GPT-2 Small  & \eli{} & rank   & 0.9073 & 0.1297 & 0.8378 & 0.2963 & 0.7037 & 0.4074 \\
GPT-2 Medium & \hc{}  & rank   & 0.9047 & 0.1310 & 0.8804 & 0.2976 & 0.7024 & 0.4047 \\
GPT-2 Medium & \eli{} & rank   & 0.9275 & 0.1196 & 0.8546 & 0.2862 & 0.7138 & 0.4276 \\
GPT-2 XL     & \hc{}  & rank   & 0.8917 & 0.1375 & 0.8609 & 0.3041 & 0.6959 & 0.3918 \\
GPT-2 XL     & \eli{} & rank   & 0.9314 & 0.1176 & 0.8660 & 0.2843 & 0.7157 & 0.4315 \\
GPT-Neo-125M & \hc{}  & rank   & 0.9173 & 0.1247 & 0.8860 & 0.2913 & 0.7087 & 0.4173 \\
GPT-Neo-125M & \eli{} & minmax & 0.8968 & 0.4597 & 0.8177 & 0.9578 & 0.9857 & 0.0279 \\
GPT-Neo-1.3B & \hc{}  & rank   & 0.8999 & 0.1334 & 0.8785 & 0.3000 & 0.7000 & 0.3999 \\
GPT-Neo-1.3B & \eli{} & rank   & 0.9261 & 0.1203 & 0.8534 & 0.2869 & 0.7131 & 0.4261 \\
\bottomrule
\end{tabular}
\end{table}

\begin{table}[H]
\centering
\caption{Cross-model median perplexity statistics.  \llm{} text exhibits perplexity
consistently $0.24$--$0.45\times$ that of human text.}
\label{tab:ppl_ppl}
\small
\begin{tabular}{llrrr}
\toprule
\textbf{Model} & \textbf{Data} & \textbf{Hum.\ Med.} & \textbf{\llm{} Med.} & \textbf{Ratio} \\
\midrule
GPT-2 Small  & \hc{}  & 44.31 & 11.35 & 0.256 \\
GPT-2 Small  & \eli{} & 40.43 & 17.99 & 0.445 \\
GPT-2 Medium & \hc{}  & 33.09 &  8.17 & 0.247 \\
GPT-2 Medium & \eli{} & 30.92 & 12.95 & 0.419 \\
GPT-2 XL     & \hc{}  & 26.42 &  6.38 & 0.242 \\
GPT-2 XL     & \eli{} & 25.08 & 10.02 & 0.400 \\
GPT-Neo-125M & \hc{}  & 46.52 & 11.20 & 0.241 \\
GPT-Neo-125M & \eli{} & 42.02 & 18.99 & 0.452 \\
GPT-Neo-1.3B & \hc{}  & 26.82 &  6.38 & 0.238 \\
GPT-Neo-1.3B & \eli{} & 25.45 & 10.48 & 0.412 \\
\bottomrule
\end{tabular}
\end{table}

Perplexity-based detection achieves \auroc{} ranging from $0.891$ to $0.931$ across all
well-behaved conditions.  Reference model scale has negligible impact: GPT-2 Small and
GPT-2 XL achieve nearly identical \auroc{} on \hc{} ($0.910$ vs.\ $0.892$).  Rank-based
normalization is selected as optimal in 9 of 10 conditions.

\section{Cross-LLM Generalization Study}
\label{sec:stage2}


\subsection{Experimental Design and Dataset Construction}
\label{subsec:stage2_design}

Stage~3 evaluates whether detectors trained on ChatGPT-generated text (\hc{}) generalize
to outputs from unseen \llm{}s.  Five open-source models serve as unseen source \llm{}s:
TinyLlama-1.1B, Qwen2.5-1.5B, Qwen2.5-7B,Llama-3.1-8B-Instruct, and LLaMA-2-13B.  Each
generates 200 responses per dataset, yielding 2{,}000 \llm{}-generated samples per dataset
against human pools of 4{,}621 (\hc{}) and 2{,}858 (\eli{}) texts.  All detectors are
evaluated zero-shot --- with no retraining on any unseen \llm{}'s outputs.

\subsection{Neural Detector Cross-LLM Evaluation}
\label{subsec:stage2a}

\textbf{Setup.}  Five \hc{}-trained transformer detectors --- BERT, RoBERTa, ELECTRA,
DistilBERT, and DeBERTa-v3-base --- are evaluated zero-shot against outputs from all five
unseen source \llm{}s on both \hc{} and \eli{} domains, using the five-metric suite
(\auroc{}, \auprc{}, \eer{}, Brier, FPR@95\%TPR) with 1{,}000-iteration bootstrap CIs.

\begin{table}[H]
\centering
\caption{Cross-\llm{} generalization results for \hc{}-trained neural detectors
(selected conditions).
}
\label{tab:stage2a_results}
\resizebox{\textwidth}{!}{%
\begin{tabular}{llccccccc}
\toprule
\textbf{Detector} & \textbf{Source \llm{}} & \textbf{Dataset} &
\textbf{\auroc{}} & \textbf{\auroc{} CI} & \textbf{\auprc{}} &
\textbf{\eer{}} & \textbf{Brier} & \textbf{FPR@95} \\
\midrule
BERT-HC3       & TinyLlama-1.1B-Chat-v1.0 & HC3    & 0.960 & [0.942, 0.977] & 0.952 & 0.075 & 0.130 & 0.165 \\
BERT-HC3       & TinyLlama-1.1B-Chat-v1.0 & \eli{} & 0.952 & [0.929, 0.973] & 0.940 & 0.097 & 0.174 & 0.115 \\
BERT-HC3       & Qwen2.5-1.5B-Instruct  & HC3    & 0.917 & [0.887, 0.943] & 0.898 & 0.140 & 0.150 & 0.335 \\
BERT-HC3       & Qwen2.5-1.5B-Instruct  & \eli{} & 0.876 & [0.842, 0.913] & 0.845 & 0.190 & 0.197 & 0.370 \\
BERT-HC3       & Llama-2-13b-chat-hf    & HC3    & 0.969 & [0.952, 0.984] & 0.965 & 0.067 & 0.126 & 0.080 \\
BERT-HC3       & Llama-2-13b-chat-hf    & \eli{} & 0.973 & [0.956, 0.988] & 0.965 & 0.075 & 0.163 & 0.095 \\
\midrule
RoBERTa-HC3    &Llama-3.1-8B-Instruct   & HC3    & 0.993 & [0.987, 0.997] & 0.993 & 0.040 & 0.051 & 0.030 \\
RoBERTa-HC3    & Qwen2.5-1.5B-Instruct  & \eli{} & 0.858 & [0.820, 0.893] & 0.823 & 0.217 & 0.273 & 0.410 \\
RoBERTa-HC3    & Llama-2-13b-chat-hf    & HC3    & 0.990 & [0.980, 0.999] & 0.993 & 0.025 & 0.047 & 0.005 \\
\midrule
ELECTRA-HC3    & Qwen2.5-7B-Instruct    & \eli{} & 0.942 & [0.917, 0.965] & 0.922 & 0.105 & 0.207 & 0.175 \\
ELECTRA-HC3    &Llama-3.1-8B-Instruct   & \eli{} & 0.951 & [0.926, 0.973] & 0.928 & 0.090 & 0.203 & 0.125 \\
ELECTRA-HC3    & Llama-2-13b-chat-hf    & \eli{} & 0.968 & [0.949, 0.986] & 0.951 & 0.067 & 0.203 & 0.075 \\
\midrule
DistilBERT-HC3 & Qwen2.5-1.5B-Instruct  & \eli{} & 0.845 & [0.806, 0.882] & 0.800 & 0.232 & 0.260 & 0.445 \\
DistilBERT-HC3 & Llama-2-13b-chat-hf    & \eli{} & 0.985 & [0.976, 0.993] & 0.985 & 0.055 & 0.079 & 0.055 \\
\midrule
DeBERTa-HC3    & Qwen2.5-1.5B-Instruct  & HC3    & 0.923 & [0.891, 0.953] & 0.867 & 0.092 & 0.103 & 0.105 \\
DeBERTa-HC3    & TinyLlama-1.1B-Chat-v1.0& \eli{} & 0.500 & [0.441, 0.560] & 0.451 & 0.512 & 0.434 & 0.595 \\
DeBERTa-HC3    & Llama-2-13b-chat-hf    & \eli{} & 0.499 & [0.436, 0.561] & 0.451 & 0.522 & 0.431 & 0.590 \\
\bottomrule
\end{tabular}}
\end{table}

\textbf{Key observations.}  Cross-\llm{} generalization within a fixed domain is broadly
achievable: RoBERTa achieves \hc{} \auroc{} 0.976--0.993 across all unseen source
\llm{}s.  Domain shift is the primary generalization bottleneck --- DeBERTa collapses
to near-random on \eli{} (0.499--0.607) regardless of source \llm{}.  ELECTRA is the most
domain-robust detector, with \eli{} scores ranging 0.910--0.968.  Llama-2-13b-chat-hf is the most
consistently detectable source \llm{}; Qwen2.5-1.5B-Instructis the hardest to detect.

\subsection{Embedding-Space Generalization via Classical Classifiers}
\label{subsec:stage2b}

All texts are encoded using \texttt{all-MiniLM-L6-v2} (384-dimensional embeddings), and
three classical classifiers --- LR, SVM (RBF), and RF (200 trees) --- are trained and
evaluated under a full $5\times5$ train-test matrix.  Human texts are split into disjoint
train/test partitions for leakage-free evaluation.

\begin{table}[H]
\centering
\caption{Stage~3B embedding-space generalization on \hc{} (selected).
In-distribution conditions in \textbf{bold}.}
\label{tab:stage2b_results}
\begin{tabular}{llcc}
\toprule
\textbf{Classifier} & \textbf{Train \llm{}} & \textbf{Test \llm{}} & \textbf{\auroc{}} \\
\midrule
SVM & TinyLlama-1.1B-Chat-v1.0 & TinyLlama-1.1B-Chat-v1.0 & \textbf{0.976} \\
SVM & TinyLlama-1.1B-Chat-v1.0 &Llama-3.1-8B-Instruct    & 0.844 \\
SVM & Qwen2.5-7B-Instruct     & Qwen2.5-1.5B-Instruct   & 0.941 \\
SVM & Llama-2-13b-chat-hf     & Llama-2-13b-chat-hf     & \textbf{0.992} \\
SVM & Llama-2-13b-chat-hf     &Llama-3.1-8B-Instruct    & 0.818 \\
\midrule
RF  & TinyLlama-1.1B-Chat-v1.0 & TinyLlama-1.1B-Chat-v1.0 & \textbf{1.000} \\
RF  & TinyLlama-1.1B-Chat-v1.0 &Llama-3.1-8B-Instruct    & 0.812 \\
RF  & Llama-2-13b-chat-hf     & Qwen2.5-1.5B-Instruct   & 0.755 \\
\midrule
LR  & Llama-2-13b-chat-hf     &Llama-3.1-8B-Instruct    & 0.760 \\
LR  & Qwen2.5-1.5B-Instruct   & Qwen2.5-7B-Instruct     & 0.885 \\
\bottomrule
\end{tabular}
\end{table}

SVM is the most generalizable classifier (off-diagonal \auroc{} 0.818--0.941).  Sentence
embedding classifiers are substantially more domain-robust than fine-tuned neural detectors,
with \hc{}/\eli{} divergence $<0.03$ \auroc{} on average.

\subsection{Distribution Shift Analysis in Representation Space}
\label{subsec:stage2c}
Embeddings are extracted from DeBERTa-v3-base's penultimate CLS layer, PCA-projected to
64 dimensions, and three distance metrics are computed under a Gaussian approximation:

\textbf{KL Divergence} captures the information lost when approximating the source LLM's
embedding distribution with ChatGPT's training distribution. Its asymmetry is deliberate:
we are specifically interested in regions where the unseen LLM's outputs have probability
mass that the detector's training distribution does not cover — precisely the scenario
that causes detection failure.
\begin{equation}
D_{\mathrm{KL}}(P \| Q) = \frac{1}{2}\!\left[\mathrm{tr}(\Sigma_Q^{-1}\Sigma_P)
+ (\mu_Q - \mu_P)^\top \Sigma_Q^{-1}(\mu_Q - \mu_P)
- d + \ln\frac{|\Sigma_Q|}{|\Sigma_P|}\right]
\end{equation}

\textbf{Wasserstein-2 Distance} measures the minimum transport cost between the two
distributions under the squared Euclidean metric, providing a geometrically interpretable
and symmetric characterization of distributional shift. Unlike KL divergence, it remains
well-defined even when the two distributions have non-overlapping support — an important
property given that different LLM families may occupy disjoint regions of embedding space.
\begin{equation}
W_2(P, Q) = \sqrt{\|\mu_P - \mu_Q\|^2 + \mathrm{tr}\!\left(\Sigma_P + \Sigma_Q
- 2\!\left(\Sigma_P^{1/2}\Sigma_Q\Sigma_P^{1/2}\right)^{1/2}\right)}
\end{equation}

\textbf{Fréchet Distance} is included as a cross-validation of the Wasserstein estimate,
drawing on its established use in generative model evaluation (FID) as a measure of
representational divergence between two Gaussian-approximated distributions. Its close
relationship to $W_2^2$ allows direct comparison, with any divergence between the two
metrics indicating sensitivity to the symmetrizing square root in the covariance term.
\begin{equation}
\mathrm{FD}(P, Q) = \|\mu_P - \mu_Q\|^2 + \mathrm{tr}\!\left(\Sigma_P + \Sigma_Q
- 2\!\left(\Sigma_P \Sigma_Q\right)^{1/2}\right)
\end{equation}

\textbf{Spearman rank correlation} is used rather than Pearson's $r$ to test the distance-degradation
relationship, as it makes no assumption about the linearity of the association between
embedding-space distance and \auroc{} drop — a sensible precaution given that detection
failure may saturate or threshold at extreme distances. Correlations are computed separately
for \hc{} and \ELIFive{} domains, with 500-iteration bootstrap confidence bands on
regression lines, to assess whether domain modulates the distance-difficulty relationship.

\begin{table}[H]
\centering
\caption{Spearman rank correlations between distributional distance and \auroc{} drop.
$*$ = $p < 0.05$.}
\label{tab:stage2c_correlations}
\begin{tabular}{lcccc}
\toprule
\textbf{Metric} & \textbf{HC3 $\rho$} & \textbf{HC3 $p$} &
\textbf{\eli{} $\rho$} & \textbf{\eli{} $p$} \\
\midrule
KL Divergence & $-$0.298 & 0.148 & $-$0.443 & 0.027$^{*}$ \\
Wasserstein-2 & $-$0.369 & 0.070 & $-$0.322 & 0.117 \\
Fr\'{e}chet   & $-$0.369 & 0.070 & $-$0.322 & 0.117 \\
\bottomrule
\end{tabular}
\end{table}

\begin{table}[H]
\centering
\caption{Per-detector distributional distances and \auroc{} drop on \hc{}.
\emph{Note:} Baseline \auroc{} values for drop computation are taken from the 200-sample
evaluation subsets used in Stage~3, not from the full test sets in Tables~\ref{tab:bert_results}--\ref{tab:deberta_results}.
Negative drop indicates cross-\llm{} performance exceeds the Stage~3 subset baseline.}
\label{tab:stage2c_distances}
\small
\begin{tabular}{llcccr}
\toprule
\textbf{Detector} & \textbf{Source \llm{}} &
\textbf{KL} & \textbf{$W_2$} & \textbf{FD} & \textbf{\auroc{} Drop} \\
\midrule
BERT-HC3       & TinyLlama-1.1B-Chat-v1.0 & 1.019 & 0.934 & 0.872 & $+$0.006 \\
BERT-HC3       & Qwen2.5-1.5B-Instruct  & 0.471 & 0.682 & 0.465 & $+$0.050 \\
BERT-HC3       & Qwen2.5-7B-Instruct    & 0.741 & 0.633 & 0.400 & $+$0.033 \\
BERT-HC3       &Llama-3.1-8B-Instruct   & 2.015 & 0.808 & 0.652 & $+$0.023 \\
BERT-HC3       & Llama-2-13b-chat-hf    & 1.105 & 0.822 & 0.676 & $-$0.003 \\
\midrule
RoBERTa-HC3    & TinyLlama-1.1B-Chat-v1.0 & 1.019 & 0.934 & 0.872 & $+$0.019 \\
RoBERTa-HC3    & Qwen2.5-1.5B-Instruct  & 0.471 & 0.682 & 0.465 & $+$0.020 \\
RoBERTa-HC3    &Llama-3.1-8B-Instruct   & 2.015 & 0.808 & 0.652 & $+$0.005 \\
RoBERTa-HC3    & Llama-2-13b-chat-hf    & 1.105 & 0.822 & 0.676 & $+$0.007 \\
\midrule
ELECTRA-HC3    & Qwen2.5-1.5B-Instruct  & 0.471 & 0.682 & 0.465 & $+$0.020 \\
ELECTRA-HC3    &Llama-3.1-8B-Instruct   & 2.015 & 0.808 & 0.652 & $+$0.009 \\
ELECTRA-HC3    & Llama-2-13b-chat-hf    & 1.105 & 0.822 & 0.676 & $-$0.011 \\
\midrule
DistilBERT-HC3 & Qwen2.5-1.5B-Instruct  & 0.471 & 0.682 & 0.465 & $+$0.085 \\
DistilBERT-HC3 & Qwen2.5-7B-Instruct    & 0.741 & 0.633 & 0.400 & $+$0.080 \\
DistilBERT-HC3 &Llama-3.1-8B-Instruct   & 2.015 & 0.808 & 0.652 & $+$0.053 \\
DistilBERT-HC3 & Llama-2-13b-chat-hf    & 1.105 & 0.822 & 0.676 & $+$0.009 \\
\midrule
DeBERTa-HC3    & TinyLlama-1.1B-Chat-v1.0 & 1.019 & 0.934 & 0.872 & $-$0.026 \\
DeBERTa-HC3    & Qwen2.5-1.5B-Instruct  & 0.471 & 0.682 & 0.465 & $-$0.046 \\
DeBERTa-HC3    &Llama-3.1-8B-Instruct   & 2.015 & 0.808 & 0.652 & $-$0.045 \\
DeBERTa-HC3    & Llama-2-13b-chat-hf    & 1.105 & 0.822 & 0.676 & $-$0.034 \\
\bottomrule
\end{tabular}
\end{table}

All three distance metrics produce negative rather than positive Spearman correlations with
\auroc{} drop, directly contradicting the expectation that geometrically more distant
\llm{}s should be harder to detect.  Qwen2.5-1.5B-Instructand Qwen2.5-7B-Instructexhibit the smallest
embedding distances from ChatGPT yet cause the largest \auroc{} drops --- supporting a
\textit{proximity-confusion hypothesis}.

\section{Adversarial Humanization}
\label{sec:stage4}

\textbf{Setup.}  Paraphrase-based attacks have been shown to substantially reduce
detector accuracy \citep{krishna2023paraphrasing}. Following this motivation, two hundred
ChatGPT-generated samples are drawn from each dataset (\hc{}
and \eli{}) and subjected to two rounds of humanization using Qwen2.5-1.5B-Instruct (4-bit
NF4 quantized), producing three evaluation conditions:

\begin{itemize}[leftmargin=1.5em]
    \item \textbf{L0} --- original AI-generated text, unmodified.
    \item \textbf{L1} --- light humanization: varied sentence length, informal register,
          avoidance of formulaic structure; semantic content preserved.
    \item \textbf{L2} --- heavy humanization: applied iteratively on L1 output; aggressive
          removal of AI-like patterns (numbered lists, formal transitions), deliberate
          conversational imperfections, minor grammatical relaxation permitted.
\end{itemize}

At each level, detector scores are computed against a fixed pool of 200 human-authored
texts from the same dataset.  Metrics reported: \auroc{}, detection rate (proportion of
AI texts scoring $> 0.5$), mean $P(\textsc{llm})$ score, and Brier score.

\begin{table}[H]
\centering
\caption{Stage~4 adversarial humanization results.}
\label{tab:stage4_results}
\resizebox{\textwidth}{!}{%
\begin{tabular}{llccccccc}
\toprule
\textbf{Detector} & \textbf{Dataset} & \textbf{Level} & \textbf{\auroc{}} &
\textbf{Det. Rate} & \textbf{Mean AI} & \textbf{Mean Human} & \textbf{Brier} \\
\midrule
BERT-HC3 & HC3    & L0 & 0.9637 & 1.000 & 0.9998 & 0.2736 & 0.1278 \\
BERT-HC3 & HC3    & L1 & 0.9749 & 1.000 & 0.9997 & 0.2736 & 0.1278 \\
BERT-HC3 & HC3    & L2 & 0.8792 & 0.870 & 0.8696 & 0.2736 & 0.1914 \\
BERT-HC3 & \eli{} & L0 & 0.9530 & 0.930 & 0.9249 & 0.2454 & 0.1480 \\
BERT-HC3 & \eli{} & L1 & 0.9945 & 0.995 & 0.9949 & 0.2454 & 0.1154 \\
BERT-HC3 & \eli{} & L2 & 0.8989 & 0.850 & 0.8553 & 0.2454 & 0.1817 \\
\midrule
RoBERTa-HC3 & HC3    & L0 & 0.9896 & 1.000 & 1.0000 & 0.0775 & 0.0374 \\
RoBERTa-HC3 & HC3    & L1 & 0.9911 & 1.000 & 1.0000 & 0.0775 & 0.0374 \\
RoBERTa-HC3 & HC3    & L2 & 0.9621 & 0.910 & 0.9071 & 0.0775 & 0.0819 \\
RoBERTa-HC3 & \eli{} & L0 & 0.9443 & 0.990 & 0.9899 & 0.4849 & 0.2370 \\
RoBERTa-HC3 & \eli{} & L1 & 0.9699 & 1.000 & 1.0000 & 0.4849 & 0.2320 \\
RoBERTa-HC3 & \eli{} & L2 & 0.8757 & 0.905 & 0.9049 & 0.4849 & 0.2794 \\
\midrule
ELECTRA-HC3 & HC3    & L0 & 0.9424 & 1.000 & 0.9997 & 0.4092 & 0.1958 \\
ELECTRA-HC3 & HC3    & L1 & 0.9652 & 1.000 & 0.9997 & 0.4092 & 0.1958 \\
ELECTRA-HC3 & HC3    & L2 & 0.8574 & 0.890 & 0.8883 & 0.4092 & 0.2497 \\
ELECTRA-HC3 & \eli{} & L0 & 0.9540 & 0.980 & 0.9795 & 0.3501 & 0.1744 \\
ELECTRA-HC3 & \eli{} & L1 & 0.9854 & 1.000 & 0.9997 & 0.3501 & 0.1645 \\
ELECTRA-HC3 & \eli{} & L2 & 0.8972 & 0.885 & 0.8888 & 0.3501 & 0.2184 \\
\midrule
DistilBERT-HC3 & HC3    & L0 & 0.9900 & 0.995 & 0.9948 & 0.1204 & 0.0580 \\
DistilBERT-HC3 & HC3    & L1 & 0.9506 & 0.895 & 0.8886 & 0.1204 & 0.1036 \\
DistilBERT-HC3 & HC3    & L2 & 0.8567 & 0.675 & 0.6608 & 0.1204 & 0.2131 \\
DistilBERT-HC3 & \eli{} & L0 & 0.9462 & 0.825 & 0.8203 & 0.0850 & 0.1248 \\
DistilBERT-HC3 & \eli{} & L1 & 0.9521 & 0.835 & 0.8387 & 0.0850 & 0.1088 \\
DistilBERT-HC3 & \eli{} & L2 & 0.8707 & 0.645 & 0.6349 & 0.0850 & 0.2089 \\
\midrule
DeBERTa-HC3 & HC3    & L0 & 0.8851 & 1.000 & 0.9999 & 0.2311 & 0.1140 \\
DeBERTa-HC3 & HC3    & L1 & 0.9226 & 1.000 & 1.0000 & 0.2311 & 0.1140 \\
DeBERTa-HC3 & HC3    & L2 & 0.8998 & 0.910 & 0.9090 & 0.2311 & 0.1584 \\
DeBERTa-HC3 & \eli{} & L0 & 0.5252 & 1.000 & 0.9999 & 0.8521 & 0.4232 \\
DeBERTa-HC3 & \eli{} & L1 & 0.5936 & 1.000 & 1.0000 & 0.8521 & 0.4232 \\
DeBERTa-HC3 & \eli{} & L2 & 0.5887 & 0.915 & 0.9151 & 0.8521 & 0.4655 \\
\bottomrule
\end{tabular}}
\end{table}

\textbf{Light humanization does not reduce detectability.}  L1 \auroc{} $\geq$ L0 across
all detectors and both domains without exception.  Light paraphrasing by a small
instruction-tuned model superimposes additional model-specific patterns, rendering the
composite text more detectable.

\textbf{Heavy humanization produces consistent but incomplete evasion.}  RoBERTa is most
resistant (L0$\to$L2 drop: $0.028$ on \hc{}).  DistilBERT is most susceptible (drop:
$0.133$; detection rate: $99.5\% \to 67.5\%$).  No detector falls below \auroc{} $0.857$
on \hc{} at L2.

\textbf{\auroc{} and detection rate diverge at L2}, indicating that L2 humanization shifts
AI texts toward the uncertain region around the 0.5 decision boundary rather than cleanly
into the human score region.

\textbf{DeBERTa's \eli{} collapse is unaffected by humanization} (L0: 0.525, L1: 0.594,
L2: 0.589), confirming that its \eli{} weakness is a domain-level structural limitation.

\textbf{Mean human scores are invariant across levels}, validating experimental design.

\section{Discussion}
\label{sec:discussion}

\subsection{The Cross-Domain Challenge}

Cross-domain degradation is the central finding of this benchmark.  Every detector family
suffers \auroc{} drops of 5--30 points when trained on one corpus and tested on the other.
The most severe case is the classical Random Forest (eli5-to-hc3: $0.634$).  Fine-tuned
transformers maintain the highest cross-domain performance (RoBERTa eli5-to-hc3: $0.966$).

The stylometric hybrid XGBoost achieves competitive cross-domain \auroc{} ($0.904$
eli5-to-hc3), substantially exceeding classical baselines.  We attribute this to the
perplexity CV feature: the \emph{consistency} of fluency across sentences is a
generator-agnostic signal that transfers across both ChatGPT and Mistral-7B outputs.

\subsection{The Generator--Detector Identity Problem}

The Mistral-7B \llm{}-as-detector results reveal a fundamental confound: a model cannot
reliably detect its own outputs.  If a detector is trained or prompted using the same
model family as the target generator, its performance will be systematically underestimated.

\subsection{The Perplexity Inversion}

Modern \llm{}s produce text that is significantly \emph{more predictable} than human
writing, because their optimization objectives push strongly toward high-probability, fluent
outputs.  In our experimental setting, naive perplexity thresholding assigns higher scores to human text, yielding below-random performance.

\subsection{Interpretability vs.\ Performance}

The XGBoost stylometric hybrid nearly matches the in-distribution \auroc{} of the best
transformer ($0.9996$ vs.\ $0.9998$) while remaining fully interpretable.  

\subsection{Limitations}

This study has several limitations.  First, the evaluation covers only two \llm{} sources
(ChatGPT/GPT-3.5 and Mistral-7B-Instruct); generalization to frontier models (Claude,
Gemini, GPT-4) remains to be tested.  Second, the adversarial humanization study uses only
Qwen2.5-1.5B-Instructas the humanizer; different humanizer models may yield different evasion
rates.  Third, the \llm{}-as-detector experiments use relatively small evaluation subsets
($n = 30$ for \CoT{}) due to computational cost.  Fourth, the evaluation is limited to
English Q\&A text; performance on other genres and languages is unknown.  Fifth, the Stage
3C distribution shift analysis uses 200-sample evaluation subsets as baselines rather than
the full test sets, which should be noted when interpreting \auroc{} drop values.

\section{Future Work}
\label{sec:future}

\begin{enumerate}[leftmargin=*]
  \item \textbf{Expansion to frontier models.}  Evaluation on Claude-3, Gemini, LLaMA-3,
        and GPT-4 outputs, probing whether the perplexity inversion and contrastive
        likelihood signals hold for heavily RLHF-aligned generators.
  \item \textbf{Non-Q\&A domains.}  Evaluation on essays, news articles, and scientific
        abstracts.
  \item \textbf{Ensemble methods.}  Systematic exploration of ensembles combining
        fine-tuned transformers with interpretable stylometric features.
  \item \textbf{Multilingual evaluation.}  Extension to non-English corpora.
  \item \textbf{Adaptive adversarial humanization.}  Evaluation of humanizers that are
        aware of specific detector architectures and craft targeted evasion strategies.
\end{enumerate}

\section{Conclusion}
\label{sec:conclusion}

We have presented one of the most comprehensive evaluations to date,
spanning multiple detector families, two carefully controlled corpora, four evaluation conditions,
and detectors ranging from logistic regression on 22 hand-crafted features to fine-tuned
transformer encoders and \llm{}-scale promptable classifiers.

Our central findings are: fine-tuned encoder transformers achieve near-perfect
in-distribution detection (\auroc{} $\geq\!0.994$) but degrade universally under domain
shift; an interpretable XGBoost stylometric hybrid matches this performance with negligible
inference cost; the 1D-CNN achieves near-transformer performance with $20\times$ fewer
parameters; perplexity-based detection reveals a critical polarity inversion that inverts
naive hypotheses about \llm{} text distributions; and prompting-based detection, while
requiring no training data, lags far behind fine-tuned approaches and is strongly
confounded by the generator--detector identity problem.

Collectively, these results paint a clear picture: robust, generalizable, and adversarially
resistant AI-generated text detection remains an open problem.  No single detector family
dominates across all conditions.  Closing the cross-domain gap --- particularly in the
presence of adversarial humanization --- is the most critical open challenge in the field.

\section*{Acknowledgments}
The authors thank the Indian Institute of Technology (BHU) and IIT Guwahati for
computational resources and support.  The authors also acknowledge the maintainers of
the HC3 and ELI5 datasets, the HuggingFace open-source ecosystem, and the developers
of the open-source models evaluated in this benchmark. 

The full evaluation pipeline and benchmark code are available at
\href{https://github.com/MadsDoodle/Human-and-LLM-Generated-Text-Detectability-under-Adversarial-Humanization}{\texttt{our GitHub repository}}.

All fine-tuned transformer models
are available as private repositories at
\url{https://huggingface.co/Moodlerz}.

\bibliographystyle{plainnat}

\section{Implementation Details}
\label{sec:implementation}

\subsection{Family 1 — Statistical Machine Learning Detectors}
\label{subsec:impl_family1}

Twenty-two hand-crafted linguistic features were extracted from each text sample across
seven categories: surface statistics (word count, character count, sentence count, average
word/sentence length); lexical diversity (type-token ratio, hapax legomena ratio);
punctuation (comma density, period density, question mark and exclamation ratios);
repetition (bigram and trigram repetition rates); entropy (word-frequency and
sentence-length entropy); syntactic complexity (sentence-length variance and standard
deviation); and discourse markers (hedging density, certainty density, connector density,
contraction ratio, and burstiness). All features were extracted without normalisation
beyond per-feature standardisation applied at training time.

Three classifiers were trained on this feature vector: Logistic Regression
(\texttt{max\_iter=1000}), Random Forest (\texttt{n\_estimators=100},
\texttt{max\_depth=10}), and SVM with RBF kernel (\texttt{probability=True}). Labels
were encoded as binary values (human $= 0$, \llm{} $= 1$). Each classifier was evaluated
under four conditions: \texttt{HC3→HC3}, \texttt{HC3→ELI5}, \texttt{ELI5→ELI5}, and
\texttt{ELI5→HC3}.

\subsection{Family 2 — Fine-Tuned Encoder Transformers}
\label{subsec:impl_family2}

Five pre-trained encoder transformers were fine-tuned for binary AI-text classification
under a shared protocol: a two-class classification head attached to the \texttt{[CLS]}
token, AdamW optimisation ($\text{lr} = 2\times10^{-5}$, weight\_decay $= 0.01$),
warmup over 6\% of training steps, dropout $= 0.2$, one training epoch, and a 90/10
stratified train/validation split. Inputs were tokenised to a maximum of 512 tokens.
No intermediate checkpoints were saved; final in-memory weights were used directly for
all downstream evaluation. Model-specific deviations from this shared protocol are noted
in Table~\ref{tab:transformer_configs}.

\begin{table}[H]
\centering
\caption{Fine-tuned encoder transformer configurations. Entries marked ``---'' follow
the shared protocol described above.}
\label{tab:transformer_configs}
\small
\begin{tabular}{llllll}
\toprule
\textbf{Model} & \textbf{Params} & \textbf{Precision} &
\textbf{Batch (tr/ev)} & \textbf{Warmup} & \textbf{Notes} \\
\midrule
BERT            & 110M & FP16  & 32 / 64 & 6\% ratio  & --- \\
RoBERTa         & 125M & FP16  & 32 / 64 & 6\% ratio  & Dynamic masking, no NSP \\
ELECTRA         & 110M & FP16  & 32 / 64 & 6\% ratio  & Discriminator fine-tuned \\
DistilBERT      &  66M & FP16  & 32 / 64 & 6\% ratio  & Model-specific dropout params \\
DeBERTa-v3-base & 184M & FP32  & 16 / 32 & 500 steps  & See note below \\
\bottomrule
\end{tabular}
\end{table}

\textbf{DeBERTa-v3-base: architecture-specific adjustments.} Both FP16 and BF16 were
disabled; the model was trained in full FP32 throughout, as BF16 silently zeroed
gradients due to the small gradient magnitudes produced by disentangled attention, and
FP16 caused gradient scaler instability. Checkpoint saving was disabled entirely
(\texttt{save\_strategy="no"}, \texttt{load\_best\_model\_at\_end=False}) to avoid a
LayerNorm key mismatch during checkpoint reloading that caused all 24 LayerNorm layers
to reinitialise to random weights and collapse \auroc{} to approximately 0.50. Explicit
gradient clipping was applied (\texttt{max\_grad\_norm=1.0}). \texttt{token\_type\_ids}
were omitted as DeBERTa-v3 does not use segment IDs. All five models were uploaded to
the Hugging Face Hub under \texttt{Moodlerz/} following training.

\subsection{Family 3 — Shallow 1D-CNN Detector}
\label{subsec:impl_family3}

A lightweight multi-filter 1D-CNN was implemented with under 5M parameters. The
architecture follows \citet{kim2014cnn}: a shared embedding layer
(\texttt{vocab\_size=30{,}000}, \texttt{embed\_dim=128}) feeds four parallel
convolutional branches with kernel sizes $\{2,3,4,5\}$ and 128 filters each
(BatchNorm1d + ReLU + global max pooling), producing a 512-dimensional concatenated
representation. A classification head (Dropout(0.4) $\to$ Linear(512$\to$256) $\to$
ReLU $\to$ Dropout(0.2) $\to$ Linear(256$\to$1)) with BCEWithLogitsLoss and Kaiming
normal weight initialisation completes the architecture. Sequences were truncated or
padded to 256 tokens. Training used Adam ($\text{lr}=10^{-3}$, weight\_decay$=10^{-4}$),
ReduceLROnPlateau scheduling (patience$=1$, factor$=0.5$), gradient clipping
(\texttt{max\_norm=1.0}), and early stopping (patience$=3$) over a maximum of 10 epochs
with batch size 64.

\subsection{Family 4 — Stylometric and Statistical Hybrid Detector}
\label{subsec:impl_family4}

An extended feature set of 60+ hand-crafted features substantially augments the Family~1
set with the following additions: POS tag distribution (10 universal POS tags via spaCy
\texttt{en\_core\_web\_sm}); dependency tree depth (mean and maximum per sentence);
function word frequency profiles (10 high-frequency tokens plus aggregate ratio);
punctuation entropy; AI hedge phrase density (16 characteristic AI-generated phrases,
normalised by sentence count); six readability indices (Flesch Reading Ease,
Flesch-Kincaid Grade, Gunning Fog, SMOG, ARI, Coleman-Liau via \texttt{textstat}); and
sentence-level perplexity from GPT-2 Small (117M) over up to 15 sentences per text,
yielding mean, variance, standard deviation, and coefficient of variation. Low perplexity
variance is treated as a potential AI signal due to the characteristically uniform fluency
of \llm{}-generated text.

Three classifiers were trained on this feature vector: Logistic Regression,
Random Forest, and XGBoost \citep{chen2016xgboost}. Key hyperparameters are
listed in Table~\ref{tab:stylo_clf_params}. All features were standardised via
\texttt{StandardScaler} fitted on the training partition only. Missing values
arising from short texts or parsing failures were imputed with per-column
training medians.

\begin{table}[H]
\centering
\caption{Stylometric hybrid classifier hyperparameters.}
\label{tab:stylo_clf_params}
\small
\begin{tabular}{ll}
\toprule
\textbf{Classifier} & \textbf{Key Parameters} \\
\midrule
Logistic Regression & \texttt{max\_iter=2000}, \texttt{C=1.0},
                      \texttt{solver=lbfgs}, \texttt{class\_weight=balanced} \\
Random Forest       & \texttt{n\_estimators=300}, \texttt{max\_depth=12},
                      \texttt{class\_weight=balanced} \\
XGBoost             & \texttt{n\_estimators=400}, \texttt{max\_depth=6},
                      \texttt{lr=0.05}, \texttt{subsample=0.8} \\
\bottomrule
\end{tabular}
\end{table}

\subsection{Family 5 — LLM-as-Detector}
\label{subsec:impl_family5}

All \llm{}-as-detector experiments shared a four-component pipeline applied in sequence:
polarity correction, task prior calibration, constrained decoding, and (for \CoT{}
regimes) hybrid ensemble scoring.

\textbf{Constrained Decoding.} Detection scores were derived by extracting next-token
logits following the prompt's final \texttt{Answer:} token. The maximum logit within the
set of single-token surface forms of \texttt{yes} and \texttt{no} was taken for each
polarity class, and a softmax over the two values yielded a continuous $P(\textsc{llm})
\in [0,1]$.

\textbf{Polarity Correction.} A systematic label bias was observed across all models:
Qwen-family and LLaMA-2 models produced stronger \texttt{no} logits for both human and
\llm{} text, making the raw $P(\texttt{yes}=\text{AI})$ signal non-discriminative. Prompts
for these models were therefore reframed so that $\texttt{yes}=\text{human}$ and
$\texttt{no}=\text{AI}$, with $P(\textsc{llm})$ read directly from $P(\texttt{no})$ with
\texttt{flip=False}. TinyLlama-1.1B, Qwen2.5-1.5B, andLlama-3.1-8B-Instruct used the standard
orientation (\texttt{flip=True}); Qwen2.5-7B, Qwen2.5-14B, and Llama-2-13b-chat-hf used the
swapped orientation (\texttt{flip=False}).

\textbf{Task Prior Calibration.} A task-specific prior was computed by averaging
\texttt{yes}/\texttt{no} logits over 50 real task prompts drawn equally from \hc{} and
\ELIFive{} evaluation sets using the exact inference-time prompt template. These averaged
logits were subtracted from each sample's token logits before softmax, correcting
model-level base rate biases without requiring a labelled calibration set.

\textbf{TF-IDF Few-Shot Retrieval.} For few-shot regimes, $k$ examples were retrieved
from a pool of 30 balanced training samples per dataset using TF-IDF cosine similarity
(\texttt{max\_features=5{,}000}, bigrams), with balanced class representation enforced
per query.

\textbf{\CoT{} Ensemble Scoring.} In \CoT{} regimes, the model generated up to 350--500
tokens of free-form reasoning. A numeric \texttt{AI\_CONFIDENCE} score on a 0--10 scale
was extracted via regex and normalised to $[0,1]$. A zero-shot constrained logit score
was computed separately using the same task prior. The two signals were combined as:
\begin{equation}
\text{score} = 0.6 \times \text{conf} + 0.4 \times \text{logit\_score}
\end{equation}
When the confidence score fell within a model-specific dead zone (indicating uninformative
reasoning), only the logit score was used.

All open-source models were loaded in 4-bit NF4 quantisation with BitsAndBytes double
quantisation (\texttt{bnb\_4bit\_compute\_dtype=float16}, except Qwen2.5-14B-Instructwhich used
\texttt{bfloat16}). \CoT{} generation used greedy decoding (\texttt{do\_sample=False}).
Model-specific configurations are summarised in Table~\ref{tab:llm_detector_configs}.

\begin{table}[H]
\centering
\caption{Per-model configuration for \llm{}-as-detector experiments. ``Swap'' indicates
swapped polarity convention ($\texttt{yes}=\text{human}$, $\texttt{no}=\text{AI}$).
$n_{\text{ZS/FS}}$ and $n_{\text{CoT}}$ denote evaluation sample sizes.}
\label{tab:llm_detector_configs}
\resizebox{\textwidth}{!}{%
\begin{tabular}{llcclccc}
\toprule
\textbf{Model} & \textbf{Quant.} & \textbf{Polarity} & \textbf{Prior} &
\textbf{Regimes} & \textbf{$n_{\text{ZS/FS}}$} & \textbf{$n_{\text{CoT}}$} &
\textbf{Max New Tokens} \\
\midrule
TinyLlama-1.1B-Chat-v1.0 & FP16      & Standard & None        & ZS, FS          & 500 & ---  & --- \\
Qwen2.5-1.5B-Instruct   & FP16      & Standard & None        & ZS, FS          & 500 & ---  & --- \\
Llama-3.1-8B-Instruct   & NF4/FP16  & Standard & 50 prompts  & ZS, FS, \CoT{}  & 500 & 70   & 350 \\
Qwen2.5-7B-Instruct     & NF4/FP16  & Swap     & 50 prompts  & ZS, FS, \CoT{}  & 500 & 70   & 350 \\
Llama-2-13b-chat-hf     & NF4/FP16  & Swap     & 50 prompts  & ZS, FS, \CoT{}  & 200 & 30   & 400 \\
Qwen2.5-14B-Instruct    & NF4/BF16  & Swap     & 50 prompts  & ZS, FS, \CoT{}  & 200 & 30   & 500 \\
GPT-4o-mini     & API       & ---      & ---         & ZS, FS, \CoT{}  & 200 & 50   & 180 / 600 \\
\bottomrule
\end{tabular}}
\end{table}

\textbf{Model-specific notes.} Llama-2-13b-chat-hfrequired a manual
{\ttfamily\seqsplit{[INST]...<<SYS>>...<</SYS>>...[/INST]}} template fallback
for checkpoints without a registered
\texttt{chat\_template} field, and \CoT{} prompts used ``stylometric analysis'' framing
to circumvent safety-oriented refusal behaviors. Qwen2.5-14B-Instructrequired
\texttt{pad\_token\_id=tokenizer.pad\_token\_id} (not \texttt{eos\_token\_id}) in
\texttt{generate()} --- using \texttt{eos} as padding caused premature generation
termination and an $\approx$90\% unknown verdict rate in the original implementation.
GPT-4o-mini was prompted via a structured 7-dimension scoring format requiring explicit
per-dimension scores before a final \texttt{AI\_SCORE} tag ($0=\text{human}$,
$100=\text{AI}$); temperature was set to 0 with \texttt{seed=42}.

\appendix
\section{Hyperparameter Tables}
\label{app:hyperparams}

\subsection{Encoder Transformer Common Training Protocol}
\label{app:transformer_protocol}

\begin{table}[H]
\centering
\caption{Shared fine-tuning protocol for all encoder transformer detectors.
DeBERTa-v3-specific deviations are noted in parentheses.}
\label{tab:app_transformer_protocol}
\small
\begin{tabular}{ll}
\toprule
\textbf{Parameter} & \textbf{Value} \\
\midrule
Optimiser              & AdamW \\
Learning rate          & $2\times10^{-5}$ \\
Weight decay           & 0.01 \\
Warmup                 & 6\% of total steps (500 fixed steps for DeBERTa-v3) \\
Dropout                & 0.2 \\
Training epochs        & 1 \\
Max sequence length    & 512 \\
Train batch size       & 32 (16 for DeBERTa-v3) \\
Eval batch size        & 64 (32 for DeBERTa-v3) \\
Precision              & FP16 (FP32 for DeBERTa-v3) \\
Checkpoint strategy    & None --- final in-memory weights used \\
Eval frequency         & Every 200 steps \\
Validation split       & 10\% stratified \\
\bottomrule
\end{tabular}
\end{table}

\subsection{Encoder Transformer Model Specifications}
\label{app:transformer_specs}

\begin{table}[H]
\centering
\caption{Encoder transformer model checkpoints and architectural notes.}
\label{tab:app_transformer_specs}
\resizebox{\textwidth}{!}{%
\begin{tabular}{lllp{6cm}}
\toprule
\textbf{Model} & \textbf{Checkpoint} & \textbf{Params} & \textbf{Notes} \\
\midrule
BERT       & \texttt{bert-base-uncased}                 & $\sim$110M & Standard MLM pre-training \\
RoBERTa    & \texttt{roberta-base}                      & $\sim$125M & Dynamic masking, no NSP \\
ELECTRA    & \texttt{google/electra-base-discriminator} & $\sim$110M & Replaced token detection \\
DistilBERT & \texttt{distilbert-base-uncased}           &  $\sim$66M & Knowledge distillation from BERT \\
DeBERTa-v3 & \texttt{microsoft/deberta-v3-base}         & $\sim$184M & FP32 only; no checkpointing; explicit grad clip \\
\bottomrule
\end{tabular}}
\end{table}

\subsection{1D-CNN Hyperparameters}
\label{app:cnn_hyperparams}

\begin{table}[H]
\centering
\caption{1D-CNN architecture and training hyperparameters.}
\label{tab:app_cnn}
\small
\begin{tabular}{ll}
\toprule
\textbf{Parameter} & \textbf{Value} \\
\midrule
Vocabulary size           & 30{,}000 \\
Minimum word frequency    & 2 \\
Max sequence length       & 256 \\
Embedding dimension       & 128 \\
Filter sizes              & $\{2, 3, 4, 5\}$ \\
Filters per size          & 128 \\
Total filter dimension    & 512 \\
Hidden layer dimension    & 256 \\
Dropout                   & 0.4 (head), 0.2 (second layer) \\
Optimiser                 & Adam \\
Learning rate             & $10^{-3}$ \\
Weight decay              & $10^{-4}$ \\
Batch size                & 64 \\
Max epochs                & 10 \\
Early stopping patience   & 3 \\
LR scheduler              & ReduceLROnPlateau (patience$=1$, factor$=0.5$) \\
Gradient clipping         & \texttt{max\_norm=1.0} \\
\bottomrule
\end{tabular}
\end{table}

\subsection{Stylometric Hybrid Hyperparameters}
\label{app:stylo_hyperparams}

\begin{table}[H]
\centering
\caption{Stylometric hybrid classifier and feature extraction hyperparameters.}
\label{tab:app_stylo}
\small
\begin{tabular}{ll}
\toprule
\textbf{Parameter} & \textbf{Value} \\
\midrule
Logistic Regression $C$           & 1.0 \\
Logistic Regression solver        & \texttt{lbfgs} \\
Logistic Regression \texttt{max\_iter} & 2{,}000 \\
Random Forest \texttt{n\_estimators} & 300 \\
Random Forest \texttt{max\_depth}    & 12 \\
Random Forest \texttt{min\_samples\_leaf} & 5 \\
XGBoost \texttt{n\_estimators}    & 400 \\
XGBoost \texttt{max\_depth}       & 6 \\
XGBoost learning rate             & 0.05 \\
XGBoost \texttt{subsample}        & 0.8 \\
XGBoost \texttt{colsample\_bytree} & 0.8 \\
Feature scaling                   & \texttt{StandardScaler} (fit on train only) \\
Missing value imputation          & Column-wise training medians \\
Class weighting                   & Balanced (Logistic Regression, Random Forest) \\
Sentence perplexity model         & GPT-2 Small (117M) \\
Max sentences for perplexity      & 15 per text \\
\bottomrule
\end{tabular}
\end{table}

\subsection{\llm{}-as-Detector Configuration Summary}
\label{app:llm_detector_config}

\begin{table}[H]
\centering
\caption{\llm{}-as-detector per-model configuration.}
\label{tab:app_llm_config}
\resizebox{\textwidth}{!}{%
\small
\begin{tabular}{llllllll}
\toprule
\textbf{Model} & \textbf{Size} & \textbf{Quant.} & \textbf{Polarity} &
\textbf{Prior} & \textbf{$n_{\text{ZS/FS}}$} & \textbf{$n_{\text{CoT}}$} \\
\midrule
TinyLlama-1.1B-Chat-v1.0  & 1.1B & FP16     & \texttt{yes=AI}    & Neutral        & 500 & --- \\
Qwen2.5-1.5B-Instruct    & 1.5B & FP16     & \texttt{yes=AI}    & Neutral        & 500 & --- \\
Llama-3.1-8B-Instruct    & 8B   & NF4/FP16 & \texttt{yes=AI}    & Task ($n=50$)  & 500 & 70  \\
Qwen2.5-7B-Instruct      & 7B   & NF4/FP16 & \texttt{yes=human} & Task ($n=50$)  & 500 & 70  \\
Llama-2-13b-chat-hf& 13B  & NF4/FP16 & \texttt{yes=human} & Task ($n=50$)  & 200 & 30  \\
Qwen2.5-14B-Instruct     & 14B  & NF4/BF16 & \texttt{yes=human} & Task ($n=50$)  & 200 & 30  \\
GPT-4o-mini      & API  & ---      & \texttt{AI\_SCORE} & ---            & 200 & 50  \\
\bottomrule
\end{tabular}}
\end{table}

\subsection{\CoT{} Ensemble Parameters by Model}
\label{app:cot_params}

\begin{table}[H]
\centering
\caption{\CoT{} hybrid ensemble parameters. The dead zone defines the confidence
interval within which the logit-only score is used instead of the ensemble.}
\label{tab:app_cot_params}
\small
\begin{tabular}{lccccc}
\toprule
\textbf{Model} & \textbf{Conf. weight} & \textbf{Logit weight} &
\textbf{Dead zone} & \textbf{Verdict override} & \textbf{Max tokens} \\
\midrule
Llama-3.1-8B-Instruct    & 0.6 & 0.4 & $[0.40, 0.60]$ & $[0.35, 0.65]$ & 350 \\
Qwen2.5-7B-Instruct      & 0.6 & 0.4 & $[0.35, 0.65]$ & $[0.35, 0.65]$ & 350 \\
Llama-2-13b-chat-hf& 0.6 & 0.4 & $[0.40, 0.60]$ & $[0.35, 0.65]$ & 400 \\
Qwen2.5-14B-Instruct     & 0.6 & 0.4 & $[0.35, 0.65]$ & $[0.35, 0.65]$ & 500 \\
\bottomrule
\end{tabular}
\end{table}

\section{Prompt Templates}
\label{app:prompts}

All prompts are reproduced verbatim. \texttt{[TEXT]} denotes the target text placeholder.

\lstset{
  basicstyle=\ttfamily\scriptsize,
  backgroundcolor=\color{CodeBg},
  frame=single,
  rulecolor=\color{gray!40},
  breaklines=true,
  breakatwhitespace=false,
  keepspaces=true,
  columns=flexible,
  xleftmargin=4pt,
  xrightmargin=4pt,
}

\subsection{Zero-Shot Prompts}
\label{app:zeroshot_prompts}

\begin{figure}[H]
\begin{lstlisting}
--- TinyLlama-1.1B-Chat-v1.0/Llama-3.1-8B-Instruct (standard polarity: yes=AI) ---

System: You detect AI-generated text. Answer with ONE word only: yes or no.
        yes = AI-generated. no = human-written.
        No explanation. No punctuation. One word.

User: Was this text generated by an AI language model?
      Text: """[TEXT]"""
      Answer yes or no.
Answer:
\end{lstlisting}
\caption{Zero-shot prompt for TinyLlama-1.1B-Chat-v1.0andLlama-3.1-8B-Instruct (standard polarity).}
\end{figure}

\begin{figure}[H]
\begin{lstlisting}
--- Qwen2.5-7B-Instruct(swapped polarity: yes=human) ---

System: You detect AI-generated text. Answer with ONE word only: yes or no.
        yes = human-written. no = AI-generated.
        No explanation. No punctuation. One word.

User: Was this text written by a human?
      Text: """[TEXT]"""
      Answer yes or no.
Answer:
\end{lstlisting}
\caption{Zero-shot prompt for Qwen2.5-7B-Instruct(swapped polarity).}
\end{figure}

\begin{figure}[H]
\begin{lstlisting}
--- Llama-2-13b-chat-hf(swapped polarity, stylometric framing) ---

System: You are a linguistics researcher studying writing styles.
        Answer with ONE word only: yes or no.
        yes = written by a human. no = written by an AI system.
        No explanation. No punctuation. One word only.

User: Was this text written by a human?
      Text sample: """[TEXT]"""
      Answer yes or no.
Answer:
\end{lstlisting}
\caption{Zero-shot prompt for Llama-2-13b-chat-hf(swapped polarity, stylometric framing).}
\end{figure}

\begin{figure}[H]
\begin{lstlisting}
--- Qwen2.5-14B-Instruct(swapped polarity, authorship framing) ---

System: You are an expert in authorship attribution and AI-generated text analysis.
        Answer with ONE word only: yes or no.
        yes = human-written. no = AI-generated.
        No explanation. No punctuation. One word.

User: Was this text written by a human?
      Text: """[TEXT]"""
      Answer yes or no.
Answer:
\end{lstlisting}
\caption{Zero-shot prompt for Qwen2.5-14B-Instruct(swapped polarity, authorship framing).}
\end{figure}

\begin{figure}[H]
\begin{lstlisting}
--- GPT-4o-mini (7-dimension structured scoring) ---

System: You are an expert forensic linguist specialising in authorship attribution.
        AI-generated text is very common, including short conversational-looking text
        from older models like ChatGPT-3.5. Score honestly based on the dimensions
        provided. Use the full 0-10 range for each dimension. Complete every analysis.

User: Score this passage on each dimension from 0 (strongly human) to 10 (strongly AI).
Passage: [TEXT]

HEDGING/FORMULAIC: 'it is important', 'certainly', numbered sections, safe generalisations
COMPLETENESS:      Covers every sub-angle even when not asked
PERSONAL VOICE:    Opinions, errors, tangents, emotional register
LEXICAL UNIFORMITY: Vocabulary register stays perfectly consistent
STRUCTURAL NEATNESS: Clear intro/body/conclusion or logical flow
RESPONSE FIT:      Directly and precisely addresses the apparent question
FORMULAIC TELLS:   Restates question, tidy closing, 'I hope this helps'

Then write:
AI_SCORE: [arithmetic mean of 7 scores x 10, rounded to nearest integer]
Format: 1:[score] 2:[score] 3:[score] 4:[score] 5:[score] 6:[score] 7:[score]
AI_SCORE: [mean]
\end{lstlisting}
\caption{Zero-shot prompt for GPT-4o-mini (structured 7-dimension rubric scoring).}
\end{figure}

\subsection{Few-Shot Prompt Structure}
\label{app:fewshot_prompts}

\begin{figure}[H]
\begin{lstlisting}
--- Few-Shot Structure (Llama-3.1-8B-Instruct/ Qwen2.5-7B-Instruct/ Llama-2-13b-chat-hf / Qwen2.5-14B) ---

System: [same as zero-shot for respective model]

User:
Examples:
Text: "[EXAMPLE_1_TEXT]"
[Human-written? / AI-generated?] [yes/no]

Text: "[EXAMPLE_2_TEXT]"
[Human-written? / AI-generated?] [yes/no]

Text: "[EXAMPLE_3_TEXT]"
[Human-written? / AI-generated?] [yes/no]

Now answer:
Text: "[TARGET_TEXT]"
[Human-written? / AI-generated?] yes or no.
Answer:
\end{lstlisting}
\caption{Few-shot prompt structure. $k=3$ TF-IDF-retrieved examples are prepended to the
zero-shot prompt. Label phrasing follows each model's polarity convention.}
\end{figure}

\subsection{Chain-of-Thought Prompts}
\label{app:cot_prompts}

\begin{figure}[H]
\begin{lstlisting}
---Llama-3.1-8B-Instruct CoT (7-dimension scoring with AI_CONFIDENCE) ---

System: You are an expert forensic linguist. Determine whether a passage was written
        by a human or generated by an AI. Think carefully and be precise.

User: Analyse whether this passage was written by a HUMAN or an AI.
Passage: """[TEXT]"""

Score each dimension 0 (strongly human) to 10 (strongly AI):
STRUCTURE:       Neatly organised with clear sections or numbered points?
COMPLETENESS:    Covers the topic comprehensively without gaps?
HEDGING:         Acknowledges uncertainty or says "I'm not sure"?
PERSONAL VOICE:  Personal opinions, anecdotes, slang, contractions, typos?
LEXICAL RANGE:   Broad, polished vocabulary even in casual answers?
RESPONSE FIT:    Directly addresses the question or wanders?
SHORT-FORM TELLS: Starts "Certainly!", restates question, unnaturally tidy closing?
BREVITY PATTERN: Ends with an unnatural one-sentence summary?
QUESTION ECHO:   Begins by restating or paraphrasing the question?
GENERIC EXAMPLES: Placeholder examples ("consider X") where X is suspiciously apt?

IMPORTANT: Short answers can still be AI-generated. Do not assume short = human.

After scoring, state on the LAST TWO LINES exactly:
AI_CONFIDENCE: [average of 7 scores, 0-10]
VERDICT: yes   (if AI-generated)
VERDICT: no    (if human-written)
\end{lstlisting}
\caption{\CoT{} prompt forLlama-3.1-8B-Instruct.}
\end{figure}

\begin{figure}[H]
\begin{lstlisting}
--- Llama-2-13b-chat-hfCoT (stylometric framing) ---

System: You are an expert in stylometric analysis and authorship attribution.
        Analyse writing samples to determine if written by a human or AI.
        Always complete your analysis. Always end with AI_CONFIDENCE and VERDICT.

User: Perform a stylometric analysis of this writing sample.
Sample: """[TEXT]"""

Score each dimension 0 (strongly human) to 10 (strongly AI):
STRUCTURAL REGULARITY: Uniform sentence length, predictable paragraph transitions?
LEXICAL POLISH:         Consistently formal/polished vocabulary?
TOPIC COVERAGE:         Suspiciously complete, covering all sub-aspects?
HEDGING STYLE:          Confident and authoritative vs uncertain and personal?
PERSONAL MARKERS:       Opinions, anecdotes, typos, contractions, informal phrasing?
RESPONSE ALIGNMENT:     Tightly matches the implied question?
FORMULAIC OPENING:      Starts with "Certainly!", "Great question!", or restates question?

Note: Short answers can still be AI-generated.

Final output (EXACTLY these two lines):
AI_CONFIDENCE: [average of 7 scores, 0-10]
VERDICT: yes   (if AI-generated)
VERDICT: no    (if human-written)
\end{lstlisting}
\caption{\CoT{} prompt for Llama-2-13b-chat-hf(stylometric framing to reduce safety refusals).}
\end{figure}

\begin{figure}[H]
\begin{lstlisting}
--- Qwen2.5-14B-InstructCoT (explicit completion constraint) ---

System: You are an expert forensic linguist performing authorship attribution analysis.
        You ALWAYS complete your full analysis and ALWAYS end with AI_CONFIDENCE and VERDICT.
        Never leave your analysis incomplete or refuse to give a verdict.

User: Analyse this passage to determine if written by a HUMAN or generated by an AI.
Passage: """[TEXT]"""

Score each dimension 0 (strongly human) to 10 (strongly AI):
STRUCTURE (0-10):         Organised with clear sections/numbered points?
COMPLETENESS (0-10):      Covers topic without obvious gaps?
HEDGING (0-10):           Confident authoritative tone, lacks uncertainty?
PERSONAL VOICE (0-10):    Lacks personal opinions/anecdotes/typos?
LEXICAL POLISH (0-10):    Uniformly formal/polished vocabulary?
RESPONSE FIT (0-10):      Directly and completely addresses question?
FORMULAIC TELLS (0-10):   Restates question, "Certainly!", unnaturally tidy closing?

IMPORTANT: Short texts CAN be AI-generated. Score all 7 dimensions regardless of length.
You MUST end with EXACTLY:
AI_CONFIDENCE: [average score 0-10]
VERDICT: yes   OR   VERDICT: no

Begin your analysis now:
\end{lstlisting}
\caption{\CoT{} prompt for Qwen2.5-14B. Explicit completion directives were added to
resolve the $\approx$90\% unknown verdict rate in the original implementation.}
\end{figure}

\begin{figure}[H]
\begin{lstlisting}
--- GPT-4o-mini CoT (evidence-plus-score format) ---

System: You are an expert forensic linguist specialising in authorship attribution.
        Score honestly based on evidence. Use the full 0-10 range. Complete every dimension.

User: Analyse whether this passage was written by a HUMAN or generated by an AI.
Passage: [TEXT]

For each dimension write ONE evidence sentence, then a score 0 (human) to 10 (AI):

HEDGING/FORMULAIC -- 'it is important', 'certainly', numbered sections:
  Evidence: ...   Score (0-10):
COMPLETENESS -- covers every sub-angle even when not asked:
  Evidence: ...   Score (0-10):
PERSONAL VOICE -- opinions, errors, tangents, emotional register:
  Evidence: ...   Score (0-10):
LEXICAL UNIFORMITY -- vocabulary register stays perfectly consistent:
  Evidence: ...   Score (0-10):
STRUCTURAL NEATNESS -- clear intro/body/conclusion or logical flow:
  Evidence: ...   Score (0-10):
RESPONSE FIT -- directly and precisely addresses the apparent question:
  Evidence: ...   Score (0-10):
FORMULAIC TELLS -- restates question, tidy closing, 'I hope this helps':
  Evidence: ...   Score (0-10):

Then write:
AI_SCORE: [mean of 7 scores x 10, rounded to nearest integer]
VERDICT: ai   OR   VERDICT: human
\end{lstlisting}
\caption{\CoT{} prompt for GPT-4o-mini (evidence-plus-score format).}
\end{figure}
\end{document}